\documentclass[conference]{IEEEtran}
\IEEEoverridecommandlockouts
\usepackage{cite}
\usepackage{amsmath,amssymb,amsfonts}
\usepackage{algorithmic}
\usepackage{graphicx}
\usepackage{textcomp}
\usepackage{xcolor}
\usepackage[hidelinks]{hyperref}
\usepackage[normalem]{ulem}
\useunder{\uline}{\ul}{}
\def\BibTeX{{\rm B\kern-.05em{\sc i\kern-.025em b}\kern-.08em
    T\kern-.1667em\lower.7ex\hbox{E}\kern-.125emX}}
\begin{document}

\def\x{{\boldsymbol{x}}}
\def\L{{\cal L}}
\def\W{{\boldsymbol{W}}}
\def\X{{\boldsymbol{X}}}
\def\Y{{\boldsymbol{Y}}}
\def\K{{\boldsymbol{K}}}
\def\Q{{\boldsymbol{Q}}}
\def\V{{\boldsymbol{V}}}
\def\A{{\boldsymbol{A}}}
\def\tA{{\boldsymbol{\tilde{A}}}}
\def\O{{\boldsymbol{O}}}

\def\b{{\boldsymbol{b}}}
\def\y{{\boldsymbol{y}}}
\def\v{{\boldsymbol{v}}}
\def\r{{\boldsymbol{r}}}
\def\TODO{{\textcolor{red}{\bf{[TODO]}}}}

\def\bdelta{{\boldsymbol{\delta}}}

\newcommand{\norm}[1]{{\lVert #1\rVert}}
\newcommand{\comment}[1]{{\textcolor{red}{\bf{[#1]}}}}

\title{Preserving Locality in Vision Transformers for Class Incremental Learning}

\author{
	\IEEEauthorblockN{
		Bowen Zheng$^{1}$, 
		Da-Wei Zhou$^{1}$, 
		Han-Jia Ye$^{1}$ and
		De-Chuan Zhan$^{1\dagger}$\thanks{$\dagger$: Correspondence to: De-Chuan Zhan (zhandc@lamda.nju.edu.cn)}
  }
	\IEEEauthorblockA{$^{1}$National Key Laboratory for Novel Software Technology, Nanjing University
	\\  \{zhengbw, zhoudw, yehj, zhandc\}@lamda.nju.edu.cn}
}

\maketitle

\begin{abstract}
Learning new classes without forgetting is crucial for real-world applications for a classification model. Vision Transformers (ViT) recently achieve remarkable performance in Class Incremental Learning (CIL). 
Previous works mainly focus on block design and model expansion for ViTs. 
However, in this paper, we find that when the ViT is incrementally trained, the attention layers gradually lose concentration on local features.
We call this interesting phenomenon as \emph{Locality Degradation} in ViTs for CIL.
Since the low-level local information is crucial to the transferability of the representation, it is beneficial to preserve the locality in attention layers.
In this paper, we encourage the model to preserve more local information as the training procedure goes on and devise a Locality-Preserved Attention (LPA) layer to emphasize the importance of local features. 
Specifically, we incorporate the local information directly into the vanilla attention and control the initial gradients of the vanilla attention by weighting it with a small initial value. 
Extensive experiments show that the representations facilitated by LPA capture more low-level general information which is easier to transfer to follow-up tasks. 
The improved model gets consistently better performance on CIFAR100 and ImageNet100.
The source code is available at \url{https://github.com/bwnzheng/LPA_ICME2023}.
\end{abstract}


\section{Introduction}
\label{sec:intro}
Deep models are good at capturing the necessary features of images for various tasks. 
In the normal classification task, deep models refine features layer by layer to get a compact representation for each image to be distinguished by the classifier. 
However, in real-world situations, new concepts increase over time, and it is necessary to allow machine learning systems to adapt to new knowledge while keeping the previously learned knowledge. 
\emph{Class Incremental Learning} (CIL) is a scenario where new concepts incrementally emerge as new classes. 
When applied to CIL, current deep models always suffer from \emph{catastrophic forgetting}~\cite{mccloskey1989catastrophic}. 
Therefore, researchers aim to balance the model between \emph{stability} (ability to resist changes) and \emph{plasticity} (ability to adapt). 
Many models and training routines are designed to approach this goal. 
Most of them focus on convolutional architectures~\cite{8107520, hou2019learning, douillard2020podnet}. 
Recently, Vision Transformers~\cite{dosovitskiy2021an} (ViT) catch researchers' attention due to their superior performance in image classification. 
Works introducing ViT into CIL mostly focus on the block design~\cite{wang2022continual} and model expansion~\cite{douillard2022dytox}. 

However, despite the success of ViT-based models applied in CIL, we find that when the ViT is incrementally trained, the attention layers of ViTs gradually lose concentration on local features compared to the normal classification training procedure. We believe this contributes to catastrophic forgetting. 

\begin{figure}[t]
    \includegraphics[width=\linewidth]{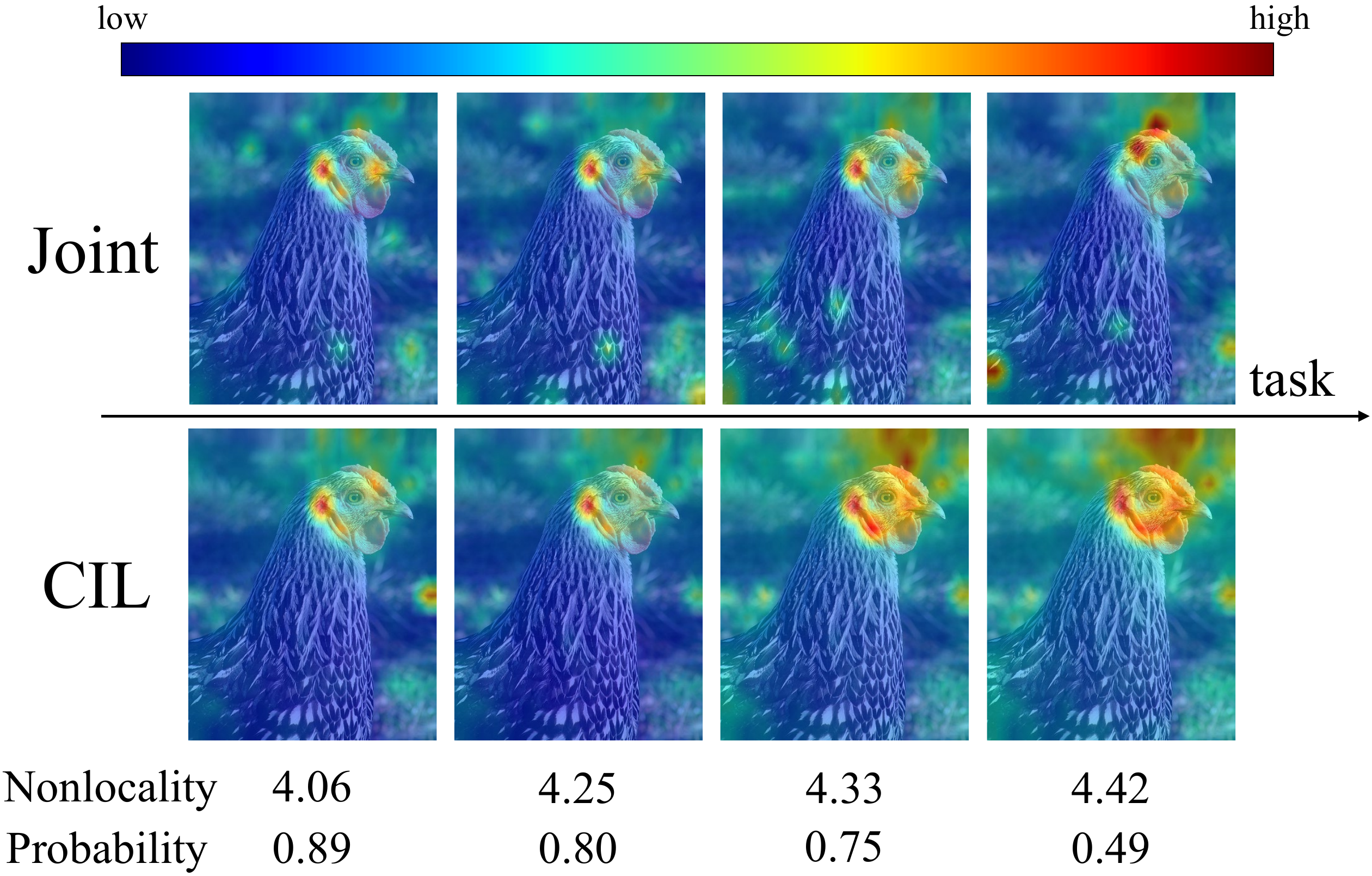}
    \caption{Visualization of Locality Degradation Phenomenon. 
    Each image is shaded by the heat map created by Attention Rollout~\cite{abnar2020quantifying}. 
    When all the presented tasks are available (Joint), the focus of the model is on the discriminative area. 
    While for CIL training, the focus spans more across the irrelevant parts (above the head), and the shaded image is warmer as the task goes on. 
    Specifically, the average nonlocality (defined in~\ref{sec:localitydegradation}) rises as the prediction probability drops. 
    More visualization results are in the appendix~\ref{sec:attnmaps}.
    }
    \label{fig:teaser}
\end{figure}
To illustrate this phenomenon, we perform a comparative experiment. 
In the experiment, we compare the nonlocality (defined in~\ref{sec:localitydegradation}) of each layer of the ViT model after the training of each task between two training procedures. 
One is the CIL training procedure, and the other is the normal classification training procedure where all of the presented classes are available for training.
The results show that, in the CIL training procedure, the nonlocality increases as the training task goes on.
That means when ViT models are trained incrementally, most of the layers will gradually focus more on global features and ignore local features. 
We call this phenomenon in CIL training as \emph{Locality Degradation}. 
For better illustration and easier understanding, we take the model after the training of each task and visualize the evolution of the heat maps using Attention Rollout~\cite{abnar2020quantifying} in Figure~\ref{fig:teaser}. 
Since it has been found that shallow layers that capture local features contain more transferable information which is task-agnostic and crucial for different tasks~\cite{yosinski2014transferable}, 
we believe locality degradation reflects the fact that ViT models are losing task-agnostic features in CIL, making the model overly focus on the current task, thus causing catastrophic forgetting. 

To alleviate locality degradation, we propose a novel attention layer called Locality-Preserved Attention (LPA) to encourage the ViT-based model to retain its ability to capture task-agnostic local features. 
Specifically, we incorporate the local information directly into the vanilla attention and control the initial gradients of the vanilla attention by weighting it with a small initial value. 
Extensive experiments show that the representations facilitated by LPA capture more low-level general information which is transferable to follow-up tasks.

Our contribution can be summarized as follows: 
\begin{itemize}
  \item We discover the locality degradation phenomenon in ViTs for CIL for the first time.
  \item We propose the Locality-Preserved Attention (LPA) layer to enhance the task-agnostic representation of the ViT-based CIL model.
  \item We perform extensive experiments to show that restricting the locality of ViTs for CIL using the LPA layer makes the representation more transferable and increases the overall accuracy.
\end{itemize}

\section{Related Works}
\label{sec:relwork}

\subsubsection{Class Incremental Learning} (CIL)~\cite{zhou2023class} is an incremental learning scenario where the model is learned task by task with a different set of classes. 
During inference, no task information of the sample is available. 
Many recent works use the model trained on previous tasks as a teacher and distillation losses to keep the previously learned knowledge~\cite{8107520, hou2019learning, douillard2020podnet}. 
LwF~\cite{8107520} proposes to use the response of the old model to guide the training of the new model's old tasks. 
PODNet~\cite{douillard2020podnet} uses the pooled intermediate feature maps of the ResNet to be the distillation target in training. 
Another popular strategy used in CIL is memory replay. 
It requires sampling a small dataset from previously learned tasks as the rehearsal samples to replay at follow-up tasks. 
The total number of rehearsal samples is fixed across the whole training procedure. 
Many works focus on how to select rehearsal samples~\cite{rebuffi2017icarl, wu2019large, tiwari2022gcr, brahma2018subset}. 
Rehearsal samples can also be obtained by generative models~\cite{shin2017continual}.
Other perspectives to boost CIL are also considered by researchers~\cite{zhou2023model, zhou2022few, shi2022mimicking, zhu2021class}. 
Reference \cite{shi2022mimicking} uses class-wise decorrelation loss to make the representation of each class at the initial task scatter more uniformly. 
Reference \cite{zhu2021class} proposes a dual augmentation framework to make the eigenvalues of the representation's covariance matrix larger. 
However, the property of the backbone network is rarely explored. 
In this work, we find the interesting locality degradation phenomenon in Vision Transformers for CIL. 
To our best knowledge, this phenomenon is discovered for the first time. 

\subsubsection{Vision Transformers} (ViT)~\cite{dosovitskiy2021an}  are designed under the inspiration of the success of fully-attention based Transformer in Natural Language Processing. 
Now ViTs have been used in various computer vision tasks. 
To remedy the disadvantages of ViT on small datasets, researchers propose to incorporate convolutional inductive bias (e.g., locality~\cite{Cordonnier2020On, d2021convit}) into ViTs. 
Convit~\cite{d2021convit} proposes to incorporate spatial information in a soft gated attention manner. 

\subsubsection{ViTs in CIL}
Works introducing ViT into CIL mainly focus on the block design~\cite{wang2022continual} and model expansion~\cite{douillard2022dytox}. 
LVT~\cite{wang2022continual} proposes to use external keys for inter-task attention. 
DyTox~\cite{douillard2022dytox} expands the task token in Convit~\cite{d2021convit} for each new task. 
Recently, researchers utilize pre-trained ViT models and adapt the model with visual prompt tuning~\cite{jia2022visual} for incremental learning~\cite{wang2022learning, wang2022sprompts, wang2022dualprompt, zhou2023revisiting}.
Due to the head start of the pre-trained models in learning representations, these methods outperform the methods which train the model from scratch, even without the rehearsal memory samples.

However, we offer a new perspective on CIL with ViTs training from scratch. 
According to our experiments, the locality is not properly preserved in CIL. 
Our work is based on this phenomenon and proposes a locality-preserved attention layer. 
The extensive experiments show that our method can help the ViT model to obtain transferable representation by preserving the locality.

\section{Methodology}
\label{sec:methodology}
\subsection{Problem Formulation}
In CIL scenarios, we have multiple tasks to learn sequentially, each task is a dataset. 
Let $D_t$ be the $t$th task. 
$(\boldsymbol{x}_i^{(t)}, y_i^{(t)})\in D_t$ is a sample. 
$\boldsymbol{x}_i^{(t)}$ is the input, $y_i^{(t)}$ is the label. 
Let $\mathcal{C}_t=\bigcup_i\{y_i^{(t)}\}$ be the class set of task $t$.
In CIL, $\forall t_1\neq t_2, \mathcal{C}_{t_1}\cap\mathcal{C}_{t_2}=\emptyset$. 
In each task, we only train the model on $D_t$, but test on all the tasks the model has trained on, i.e., the presented tasks. 
For example, when the model is training on task $t_i$, the presented tasks are tasks $t_j (j\le i)$. 
The goal is to make the model get better performance on all the presented tasks. 

The normal classification training scenario is to train on all the presented tasks from scratch rather than train the model task by task, which is a performance upper bound for CIL.
We refer to this training procedure as joint training. 

\subsection{Locality Degradation}
\label{sec:localitydegradation}
In this section, we illustrate the \emph{Locality Degradation} phenomenon in CIL. 
To start with, we introduce the \emph{nonlocality measure}~\cite{d2021convit} of a ViT layer. 
For each layer of a ViT model, suppose $\A_{ij}^{(l)(h)}$ is the attention score for patch $i$ and $j$ in layer $l$, head $h$. 
$\bdelta_{ij} = [x_j - x_i, y_j - y_i]$ is the 2-dimensional distance of patch $i$ and $j$. 
The nonlocality is defined as below:
\begin{eqnarray}
  D_{loc}^{(l)(h)} &=& \frac{1}{N}\sum_{ij}{\A_{ij}^{(l)(h)}\norm{\bdelta_{ij}}},\\
  D_{loc}^{(l)} &=& \frac{1}{N_h}\sum_{h}{D_{loc}^{(l)(h)}},
\end{eqnarray}
where $N$ is the number of patches, $N_h$ is the number of heads. 
If the layer is focusing more on local features, the nonlocality measure $D_{loc}^{(l)}$ is lower.

We perform a comparative experiment with two kinds of training procedures.
One is the CIL procedure, during the training of each task, the training set only contains the samples of the current task and limited rehearsal memory samples.
The other is the joint training procedure, where the training set contains the samples from all the presented tasks. 
We compare the nonlocality of each layer after each task during the CIL procedure with that during the joint training procedure. 
The dataset of this experiment is CIFAR100~\cite{Krizhevsky2009LearningML} and the ViT backbone is Convit~\cite{d2021convit}. 

\begin{figure}[t]
  \begin{minipage}[t]{\linewidth}
    \centering
    \includegraphics[width=\linewidth]{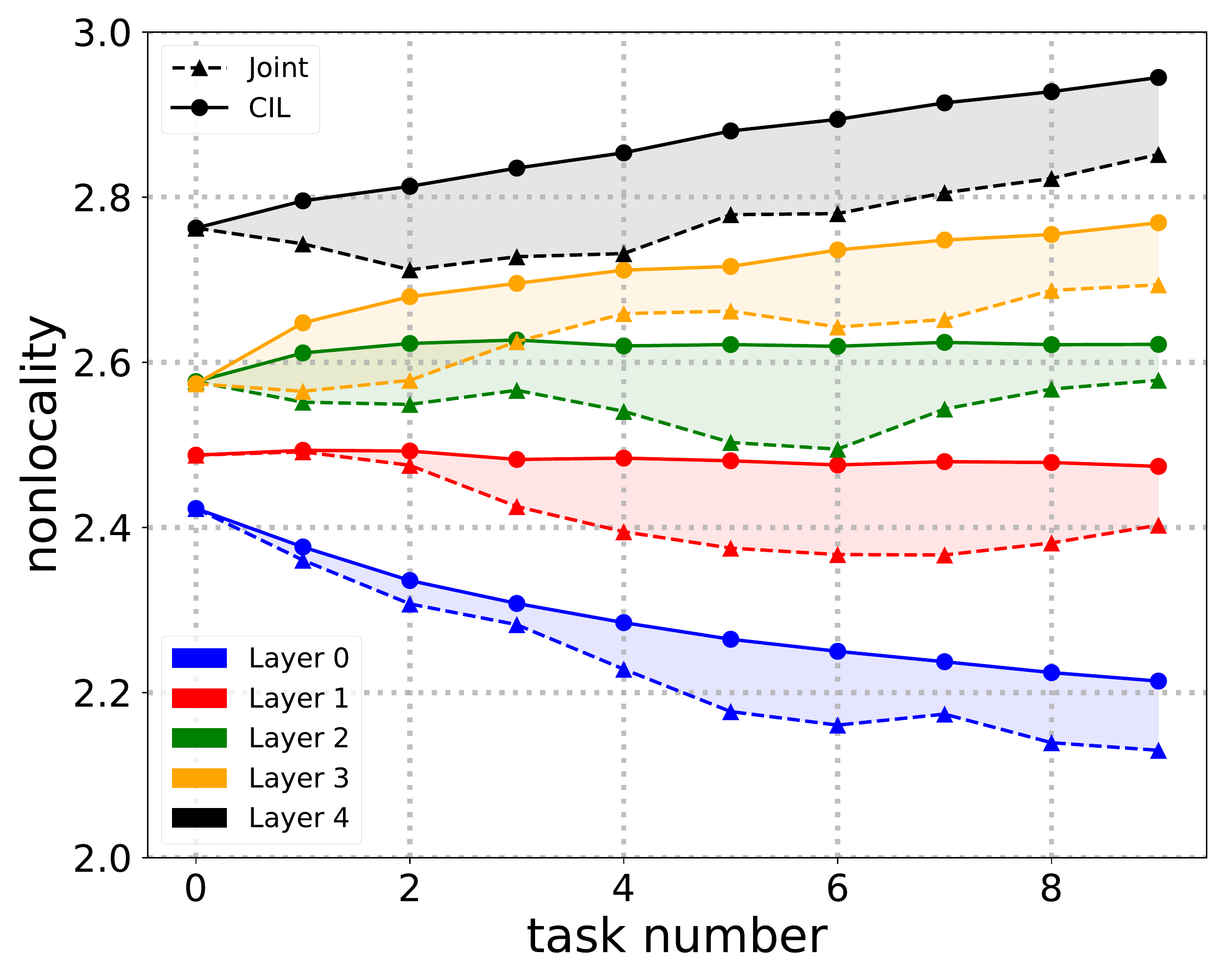}
  \end{minipage}
  \caption{The Locality Degradation Phenomenon. The CIL procedure is in the solid curves, and the joint training procedure is in the dashed curves. The colored area is the gap of nonlocality between two training procedures.}
  \label{fig:locality_degradation}
\end{figure}

The results are summarized in Figure~\ref{fig:locality_degradation}. 
It shows that (1) The nonlocality measure increases as the layers go deeper, which means shallow layers focus more on local features than deep layers. 
(2) Most of the layers, especially deep layers, are inclined to focus more on global relations as the task goes on.
(3) For joint training, the nonlocality measure is always lower than CIL at each task. 
There is always a gap in terms of the nonlocality measure between CIL and joint training procedure.
This phenomenon is what we referred to as \emph{Locality Degradation} in ViTs for CIL.

Since it has been found that shallow layers that capture local features contain more transferable information which is task-agnostic and crucial for different tasks~\cite{yosinski2014transferable},  
we believe that the nonlocality behavior of joint training is crucial for learning transferable and task-agnostic representation. 
Also, locality degradation reflects the fact that ViT models are losing task-agnostic features in CIL, making the model overly focus on the current task, thus causing catastrophic forgetting.

\begin{table*}[t]
    \caption{Performance Results on CIFAR100. Bold numbers are better between baselines and baselines with LPA. Underscored numbers are top accuracy in all the presented methods.}
    \label{tab:performance_cifar100}
    \begin{center}
        \scalebox{1.1}{
            \begin{tabular}{lccccccccccccccc}
                \hline
            Scenarios      & \multicolumn{3}{c}{B10-10}                                    & \multicolumn{1}{l}{} & \multicolumn{3}{c}{B5-5}                                      & \multicolumn{1}{l}{} & \multicolumn{3}{c}{B50-10}                                    & \multicolumn{1}{l}{} & \multicolumn{3}{c}{B50-5}                                     \\ \hline
            & Last                 & Avg                  & Fgt            & \multicolumn{1}{l}{} & Last                 & Avg                  & Fgt            & \multicolumn{1}{l}{} & Last                 & Avg                  & Fgt            & \multicolumn{1}{l}{} & Last                 & Avg                  & Fgt            \\ \hline
            iCaRL~\cite{rebuffi2017icarl}          & 50.74                & 65.27                & -              &                      & 43.75                & 61.20                & -              &                      & -                    & 65.06                & -              &                      & -                    & 58.59                & -              \\
            UCIR~\cite{hou2019learning}           & 43.39                & 58.66                & -              &                      & 40.63                & 58.17                & -              &                      & -                    & 64.28                & -              &                      & -                    & 59.92                & -              \\
            PodNet~\cite{douillard2020podnet}         & 41.05                & 58.03                & -              &                      & 35.02                & 53.97                & -              &                      & -                    & 67.25                & -              &                      & -                    & 64.04                & -              \\
            DER~\cite{yan2021dynamically}            & 65.22                & 75.36                & -              &                      & 62.48                & 74.09                & -              &                      & 65.80                & 73.21                & -              &                      & 65.21                & 72.81                & -              \\ \hline
            baseline       & 63.11                & 74.74                & 12.52          &                      & 60.23                & 72.91                & 12.47          &                      & 66.20                & 73.70                & 11.82          &                      & 63.77                & 71.86                & 14.84          \\
            baseline w/LPA & \textbf{65.27}       & \textbf{76.50}       & \textbf{11.20} &                      & \textbf{60.38}       & \textbf{73.14}       & \textbf{12.45} &                      & \textbf{68.52}       & \textbf{74.96}       & \textbf{10.02} &                      & \textbf{65.54}       & \textbf{73.08}       & \textbf{13.22} \\ \hline
            DyTox+~\cite{douillard2022dytox}         & 66.79                & 77.66                & 15.36          &                      & 62.60                & 76.02                & 21.48          &                      & 69.64                & 76.00                & \textbf{9.464} &                      & 65.70                & 73.30                & 15.13          \\
            DyTox+ w/LPA   & {\ul \textbf{68.92}} & {\ul \textbf{78.74}} & \textbf{14.30} &                      & {\ul \textbf{63.99}} & {\ul \textbf{77.21}} & \textbf{20.73} &                      & {\ul \textbf{69.76}} & {\ul \textbf{76.19}} & 9.708          &                      & {\ul \textbf{66.71}} & {\ul \textbf{74.27}} & \textbf{13.57} \\ \hline
        \end{tabular}
        }
    \end{center}
\end{table*}

\subsection{Locality-Preserved Attention Layer}
In this section, we propose the Locality-Preserved Attention Layer (LPA) to alleviate locality degradation.

The vanilla multi-head self-attention layer~\cite{dosovitskiy2021an} calculates the global similarity between each patch of the image for each head, which is formulated as below:
\begin{eqnarray}
  \Q^{(h)} &=& \W^{(h)}_{q}\X^{(h)},\\
  \K^{(h)} &=& \W^{(h)}_{k}\X^{(h)},\\
  \A^{(h)} &=& \Q^{(h)\top}\K^{(h)} / \sqrt{d_h},
\end{eqnarray}
where $h$ denotes the head number, $d_h$ is the embedding dimension of head $h$, which is usually the same for each head. 
Note that $\A^{(h)}$ is not taken softmax operation here, we will use softmax after incorporating positional information. 
We denote $\A^{(h)}_{ij}$ as the element of the matrix $\A^{(h)}$ which is the raw attention score (or patch score) for patch $i$ and $j$. 

To incorporate positional locality information into self-attention layers, we define the relative positional information between two image patches $i, j$ as following \emph{quadratic encoding}~\cite{Cordonnier2020On} formulation: 
\begin{eqnarray}
  \r_{ij} &=& \mathrm{concat}([{\norm{\bdelta_{ij}}^2}], \bdelta_{ij}),
\end{eqnarray}
$\r_{ij}$ is a 3-dimensional encoding for the positional relation between patch $i$ and $j$.

Then, we use a linear projection for each head to get the final attention score containing both patch score and positional relation encoding.
This is formulated as below, for each entry in the final attention score:
\begin{eqnarray}
  \tA^{(h)}_{ij} = \mathop{\mathrm{softmax}}_j(\mathrm{Proj}_h(\mathrm{concat}(\r_{ij}, \A^{(h)}_{ij}))),
  \label{eq:proj}
\end{eqnarray}
if we separate the parameters of the projection, Eq.~\ref{eq:proj} is equivalent to:
\begin{eqnarray}
  \tA^{(h)} = \mathop{\mathrm{softmax}}(\lambda_h\A^{(h)} + \v^{\top}_h\r),
\end{eqnarray}
where $\lambda_h$ and $\v^{\top}_h$ are learnable parameters in the projection. 
To guide the projection focusing on local features at early steps, we follow~\cite{d2021convit} and initialize $\v_h$ as the following:
\begin{eqnarray}
  \v_h = \alpha_h[-1, 2\Delta_1^{(h)}, 2\Delta_2^{(h)}],
\end{eqnarray}
where $\alpha_h$ is the locality strength hyperparameter, $\Delta^{(h)}$ is a 2-dimensional offset determining which position head $h$ should pay most attention to, relative to the query patch. For the number of heads between 9 and 15, we only initialize the first 9 heads, their $\Delta^{(h)}$ are $\Delta^{(1)}=[-1, -1]$, $\Delta^{(2)}=[-1, 0]$, $\Delta^{(3)}=[-1, 1]$, $\Delta^{(4)}=[0, -1]$, \dots, $\Delta^{(9)}=[1, 1]$.
Intuitively, it offers diversity for each head, preventing each head from producing similar results. 

For $\lambda_h$, we initialize them with a small absolute value. 
It is crucial for controlling the gradients of global attention at early training steps. And thus encourage the model to adjust its parameters mostly according to the attention with strong locality information. We will see the importance of $\lambda_h$ initialization in the ablation study.

The output of the layer follows the vanilla self-attention as below:
\begin{eqnarray}
  \V^{(h)} &=& \W^{(h)}_{v}\X^{(h)}, \\
  \O &=& \W_o\mathop{\mathrm{concat}}_{h}(\tA^{(h)}\V^{(h)}) + b_o.
\end{eqnarray}

In summary, compared to vanilla attention layers, we incorporate local information directly into the attention map.
Importantly, we initialize each $\lambda^{(h)}$ to a small absolute value to control the initial gradients of the attention, letting the parameters update according to the strong locality. 
Therefore, the model will try to classify the samples by more task-agnostic features in the first place.

\section{Experiments}
\label{sec:exps}

\subsection{Implementation Details}

To test the effectiveness of our method, we replace the vanilla attention layer with our LPA in Convit~\cite{d2021convit} as the backbone model of CIL. 
The backbone consists of 5 attention layers and 1 class attention layer. 
The class attention layer is where the class token comes in.
During CIL training, we do not expand any part of the backbone model. 

We follow \cite{douillard2022dytox} and set the maximum capacity of the replay buffer of all the experiments to be 2000. 
The base task and incremental tasks are trained for 500 epochs. 
The initial learning rate of each task is set to 0.0005 with the cosine scheduler.
The input data is augmented as standard DeiT~\cite{touvron2021training} augmentations, with Mixup~\cite{zhang2018mixup}. 

\subsection{Datasets and Baselines}
We use two common datasets to perform CIL training: CIFAR100~\cite{Krizhevsky2009LearningML} and ImageNet100~\cite{deng2009imagenet}. 
For CIL training scenarios, we apply 3 kinds of splits on classes for ImageNet100: B10-10, B50-10, B50-5, and add B5-5 for CIFAR100.
``B50-10'' means there are 50 classes in the initial task and 10 classes in each follow-up task. 

There are mainly two baselines we compare to. 
The first one is iCaRL~\cite{rebuffi2017icarl} with Convit~\cite{d2021convit} as the backbone (\emph{baseline} in Table~\ref{tab:performance_cifar100}~and~\ref{tab:performance_imagenet100}). 
In iCaRL, the nearest-mean-of-exemplars classifier is used to produce final prediction. 
The herding-based prioritized exemplar selection is used for the rehearsal memory update in each task. 
Also, it uses distillation for representation learning.
We will compare it to the backbone with LPA. 

The second one is DyTox+~\cite{douillard2022dytox}, which uses an expanding manner for class tokens. 
DyTox+ uses two-stage training within each task to finetune the classifier on the balanced rehearsal memory dataset.
Also, there are other classic and recent methods in CIL, including iCaRL~\cite{rebuffi2017icarl}, PODNet~\cite{douillard2020podnet}, DER~\cite{yan2021dynamically}, CCIL-SD~\cite{mittal2021essentials} and POD-AANets w/RMM~\cite{liu2021rmm}. 
They use ResNet18~\cite{he2016deep} for ImageNet100 and ResNet32 for CIFAR100 as the backbone.

\begin{figure}[t]
  \includegraphics[width=\linewidth]{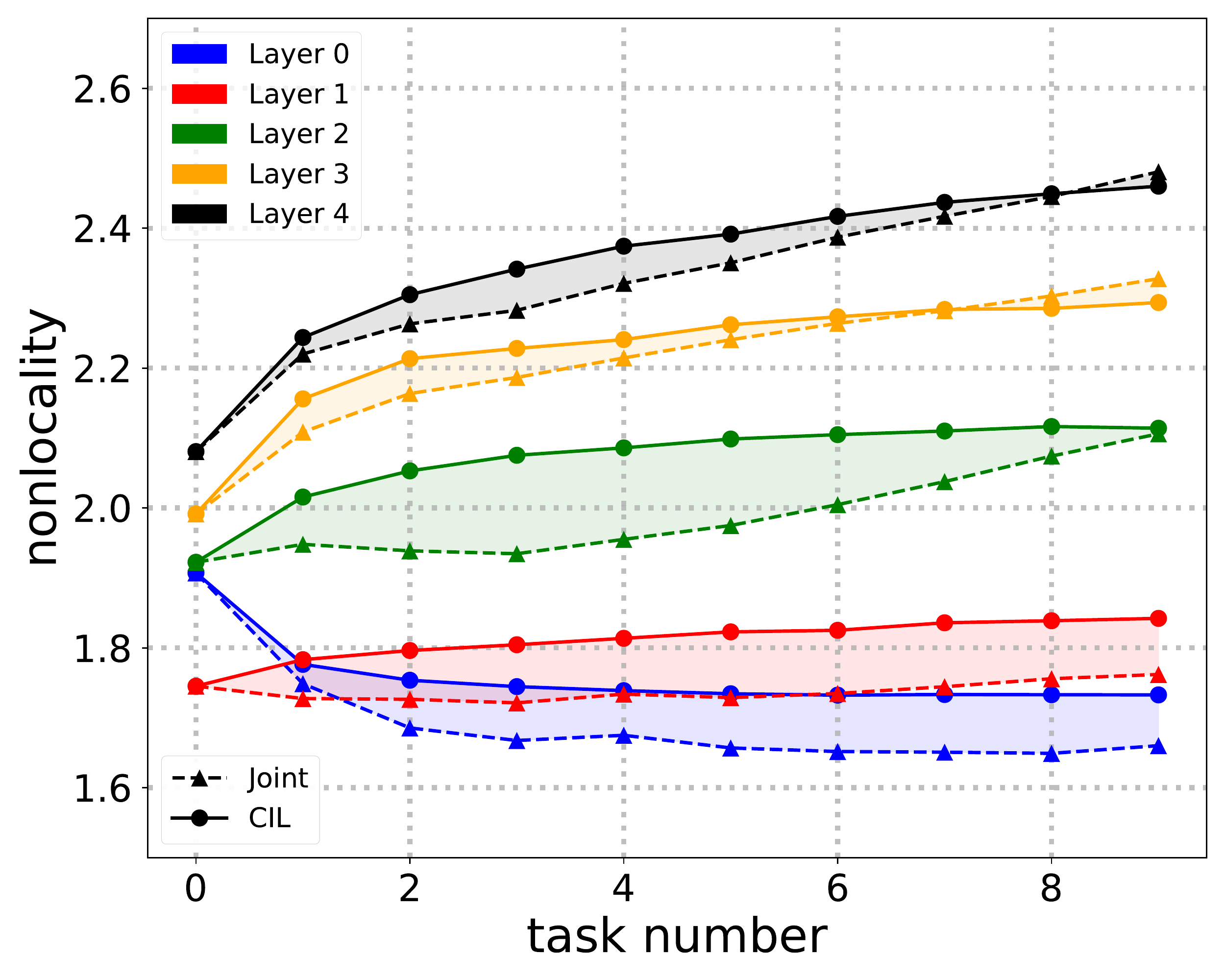}
  \caption{Locality Preserving. The CIL procedure is in the solid curves, and the joint training procedure is in the dashed curves. The colored area is the gap of nonlocality between two training procedures. }
  \label{fig:nonlocality}
\end{figure}

\begin{table*}[t]
    \caption{Performance Results on ImageNet100. Bold numbers are better between baseline and baseline with LPA. Underscored numbers are top accuracy in all the presented methods.}
    \label{tab:performance_imagenet100}
    \begin{center}
        \scalebox{1.2}{
        \begin{tabular}{lccccccccccc}
            \hline
            Scenarios      & \multicolumn{3}{c}{B10-10}                                    & \multicolumn{1}{l}{} & \multicolumn{3}{c}{B50-10}                                                                                             & \multicolumn{1}{l}{} & \multicolumn{3}{c}{B50-5}                                                                  \\ \hline
                           & Last                 & Avg                  & Fgt            & \multicolumn{1}{l}{} & Last                                  & Avg                                   & Fgt                                   & \multicolumn{1}{l}{} & Last                        & Avg                          & Fgt                          \\ \hline
            DER~\cite{yan2021dynamically}            & 66.70                & 77.18                & -              &                      & {\ul 74.92}                           & 78.20                                 & -                                     &                      & {\ul 71.64}                           & 78.42                            & -                            \\
            FOSTER~\cite{wang2022foster}                                                     & 65.68                    & 76.74                    & -              &                      & 71.60                                     & 77.37                                     & -                                     &                      & 68.20                           & 76.00                        & -                            \\
            CCIL-SD~\cite{mittal2021essentials}                                                    & -                    & -                    & -              & \multicolumn{1}{l}{} & 73.60                                     & 79.44                                 & -                                     &                      & 68.90                           & 76.77                        & -                            \\
            POD-AANets w/RMM~\cite{liu2021rmm} & -                    & -                    & -              &                      & 69.40                                     & {\ul 79.52}                           & -                                     &                      & 67.50                           & {\ul 78.47}                  & -                            \\ \hline
            baseline       & 61.02                & 72.84                & \textbf{14.38} &                      & 68.10                                 & 75.62                                 & 15.86                                 &                      & {\color[HTML]{000000} 65.4} & {\color[HTML]{000000} 74.54} & {\color[HTML]{000000} 19.75} \\
            baseline w/LPA & \textbf{61.98}       & \textbf{74.81}       & 16.16          &                      & {\color[HTML]{000000} \textbf{70.74}} & {\color[HTML]{000000} \textbf{76.83}} & {\color[HTML]{000000} \textbf{13.93}} &                      & \textbf{67.9}               & \textbf{75.38}               & \textbf{18.30}               \\ \hline
            DyTox+~\cite{douillard2022dytox}         & 65.78                & 76.35                & 17.89          &                      & 71.32                                 & 78.08                                 & \textbf{9.42}                         &                      & 66.38                       & 75.46                        & 16.92                        \\
            DyTox+ w/LPA   & {\ul \textbf{67.54}} & {\ul \textbf{77.85}} & \textbf{17.02} &                      & \textbf{71.70}                        &  \textbf{78.46}                  & 9.78                                  &                      & \textbf{68.08}        & \textbf{76.26}         & \textbf{16.31}               \\ \hline
            \end{tabular}}
    \end{center}
\end{table*}
\begin{figure*}[t]
  \begin{minipage}[t]{0.32\linewidth}
    \includegraphics[width=\linewidth]{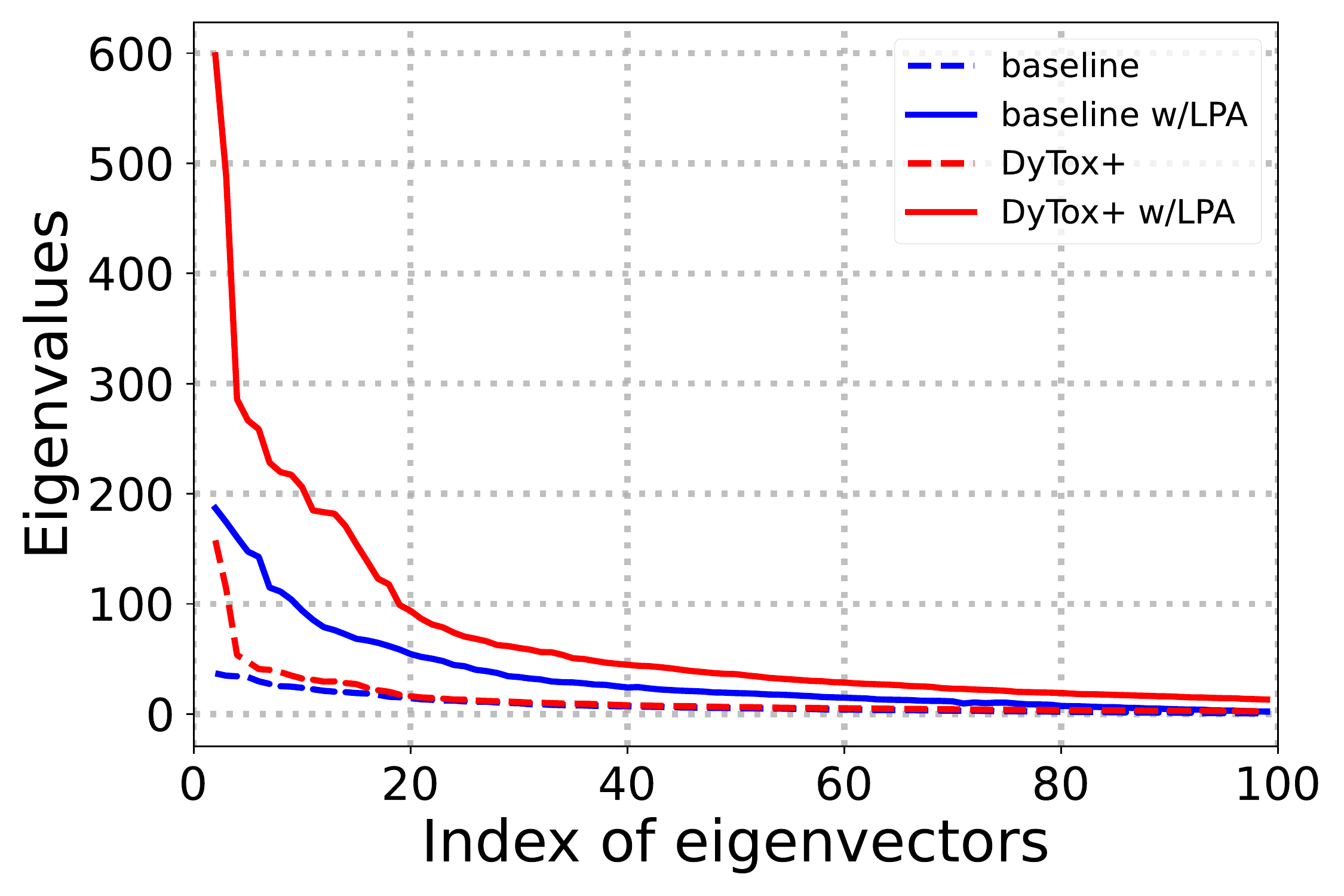}
    \caption{Eigenvalue distributions of baseline methods and with LPA. Only the 100 largest eigenvalues are presented.}
    \label{fig:eigenvalues}
  \end{minipage}
  \hfill
  \begin{minipage}[t]{0.32\linewidth}
    \includegraphics[width=\linewidth]{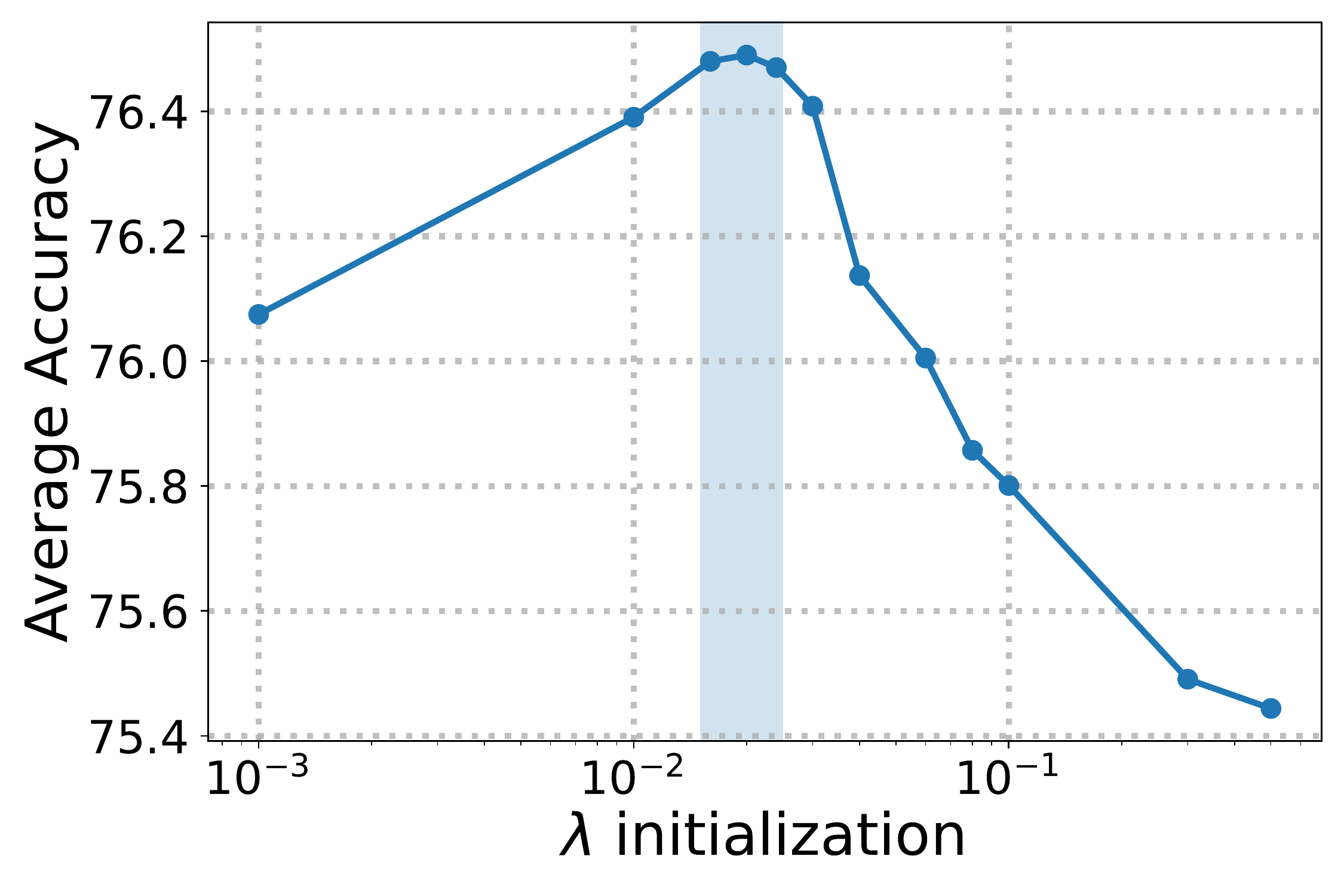}
    \caption{Ablations on $\lambda$ initialization. The shaded area is the best performance region, which is around 0.02.}
    \label{fig:lambda_abl}
  \end{minipage}
  \hfill
  \begin{minipage}[t]{0.32\linewidth}
    \includegraphics[width=\linewidth]{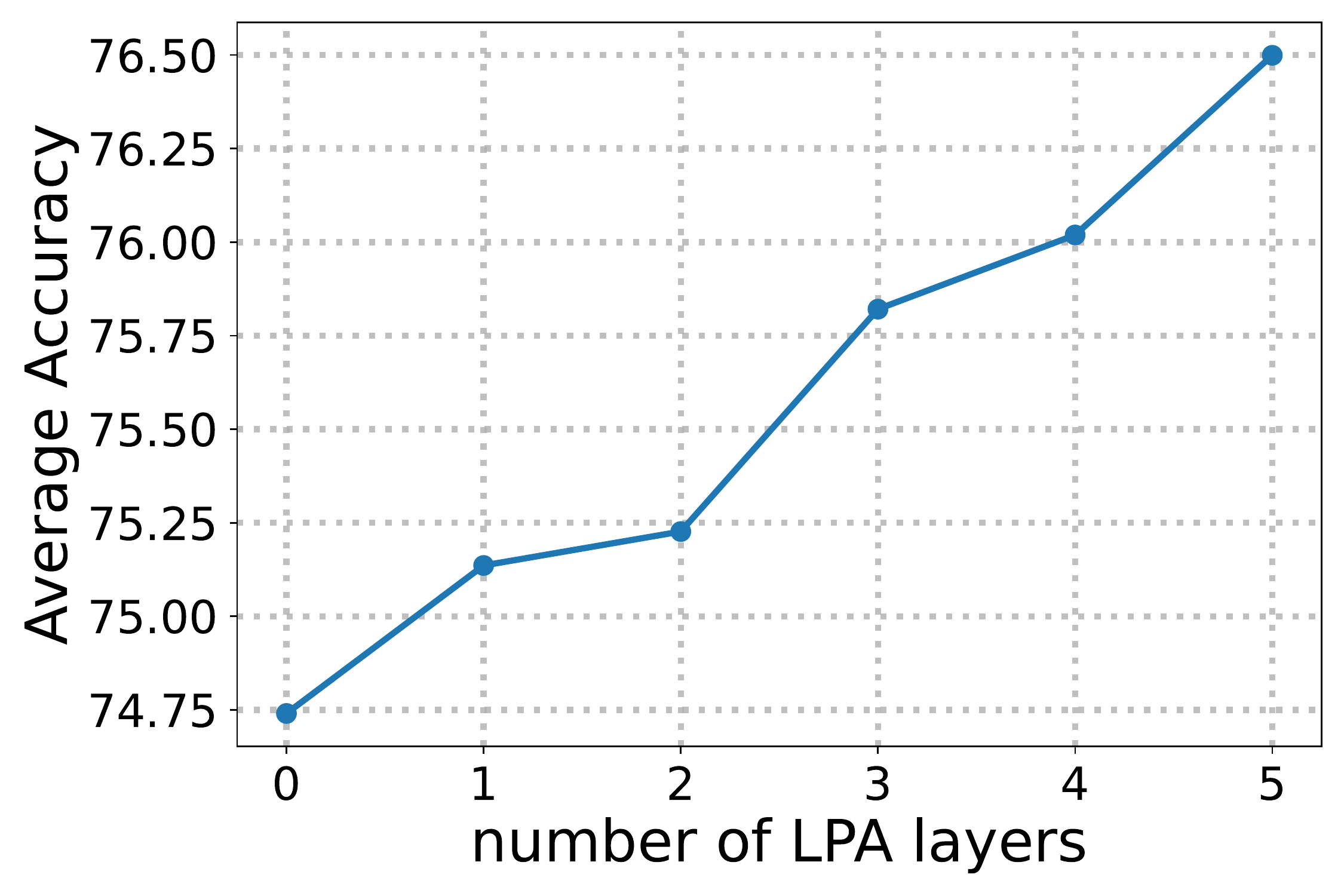}
    \caption{Ablations on the number of LPA layers. It shows more LPA layers would get better average accuracy.}
    \label{fig:lpa_layers}
  \end{minipage}
\end{figure*}

\subsection{Performance Experiments}
For the performance experiments, we report the overall top-1 accuracy of the final model on all the tasks (\emph{Last}) and average top-1 accuracy across the training procedure (\emph{Avg}). 
We also report the Forgetting~\cite{chaudhry2018riemannian} (\emph{Fgt}) of the final model. 
The Forgetting of task $k$ after task $t$ is defined as $f_t^{(k)}=\mathop{\mathrm{max}}_{i<t}{(acc_i^{(k)} - acc_t^{(k)})}, \forall t<k$, where $acc_t^{(k)}$ is the average accuracy of task $k$ in task $t$.

The results on CIFAR100 and ImageNet100 are presented in Table~\ref{tab:performance_cifar100} and Table~\ref{tab:performance_imagenet100}. 
The results of the methods other than baseline and DyTox+ are mostly taken from their papers. 
The results of the methods with LPA are averaged across 5 runs with different class orders.

The methods with LPA are generally and consistently better in each scenario on CIFAR100 and ImageNet100. 
Especially in B10-10, the baselines with LPA get around 2\% accuracy gains. 
The results show that ViT backbones with LPA gets consistently better performance in most of the CIL scenarios on two datasets.

\subsection{Locality Preserving}
To verify the effectiveness of LPA on preserving the locality, we perform a similar experiment as section~\ref{sec:localitydegradation} using the LPA-backed Convit model. 
The results are shown in Figure~\ref{fig:nonlocality}. 
As we can see, the nonlocality of CIL is growing slower as the task goes on. 
The gap between CIL and the joint procedure is smaller compared to Figure~\ref{fig:locality_degradation}.
That means the LPA layer works well on preserving the locality in the ViT model. 

\subsection{Representation Transferability}
In addition to the performance experiments, we measure the transferability of representation from another perspective. Zhu et al~\cite{zhu2021class} find that larger eigenvalues of the representation's covariance matrix transfer better and suffer less forgetting across tasks. 
This can be seen as an indicator of transferability in CIL learning.
We follow the setting of Zhu et al~\cite{zhu2021class}, comparing the eigenvalue distribution of the representation's covariance matrix. 
The representation is trained on CIFAR100 in the 10-10 scenario.
The results are shown in Figure~\ref{fig:eigenvalues}. 
As we can see, the eigenvalues are larger with LPA, which means there are more transferable directions in the representation learned with LPA. 

\subsection{Ablations}
\label{sec:ablations}
\subsubsection{Ablations on \texorpdfstring{$\lambda$}{lambda}  initialization} 
To analyze the effect of $\lambda$ initialization in LPA layers, we test the average accuracy performance of the model on CIFAR100 in the 10-10 scenario with different $\lambda$ initializations. 
We initialize each $\lambda^{(h)}$ to the same value.
The results are shown in Figure~\ref{fig:lambda_abl}. 
It is clear that the larger initialization of $\lambda$ is, the worse the performance will be, and the best performance is achieved at around 0.02.

\subsubsection{Ablations on the number of LPA layers}To analyze the effect of the number of LPA layers used in the model, we perform experiments on the intermediate models which use fewer LPA layers. 
We replace the attention layers of the baseline model from the shallowest layer one by one and test its average accuracy performance on CIFAR100 in the 10-10 scenario. 
The results are shown in Figure~\ref{fig:lpa_layers}. 
It shows that the average accuracy gets consistently improved with more LPA layers. 
The effectiveness of LPA layers is further verified.

\section{Conclusions}
\label{sec:conclusions}
In this paper, we discover an interesting phenomenon \emph{Locality Degradation} for the first time in Vision Transformers for CIL. 
This phenomenon states that when the ViT backbone is trained in the CIL procedure, the attention layers lose their ability to concentrate on local features. 
We perform comparative experiments to illustrate and verify this phenomenon in terms of the nonlocality measure. 
Since the locality is crucial for representation transferability, we then propose a \emph{Locality-Preserved Attention} layer which emphasizes and preserves the locality in the CIL training procedure. 
We incorporate the local information directly into the vanilla attention and control the initial gradients of the vanilla attention by weighting it with a small initial value.
Extensive experiments are performed to verify the effectiveness of LPA layers, and the model with LPA layers gets consistently better performance in most of the CIL scenarios.

\section*{Acknowledgment}
This work is partially supported by National Key R\&D Program of China (2022ZD0114805), NSFC (61921006, 62006112, 62250069), NSF of Jiangsu Province (BK20200313), Collaborative Innovation Center of Novel Software Technology and Industrialization.

\bibliographystyle{IEEEtran}
\bibliography{ref}

\onecolumn
\section*{Appendix}

\subsection{Attention Map Evolutions}
\label{sec:attnmaps}
\begin{figure}[!h]
    Sample Images

    \begin{minipage}[h]{0.06\linewidth}
        \includegraphics[width=\linewidth]{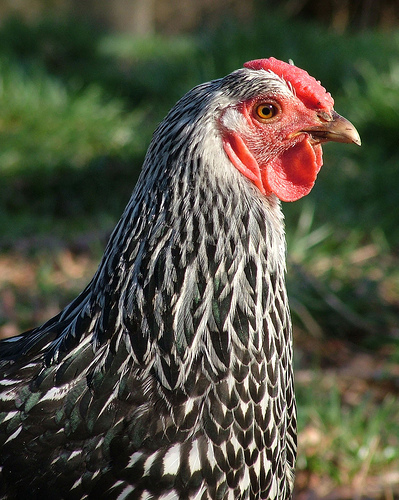}
    \end{minipage}
    \begin{minipage}[h]{0.1\linewidth}
        \includegraphics[width=\linewidth]{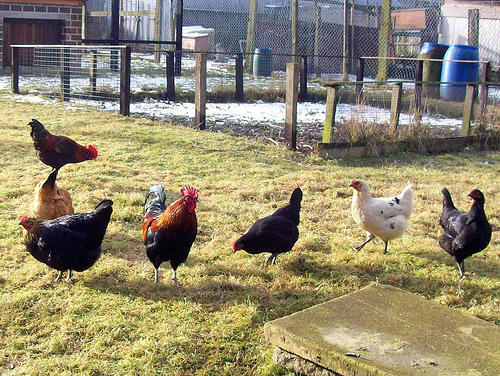}
    \end{minipage}

    \fbox{\parbox[c]{\linewidth}{
        Baseline B10-10

        \begin{minipage}[h]{0.09\linewidth}
            \includegraphics[width=\linewidth]{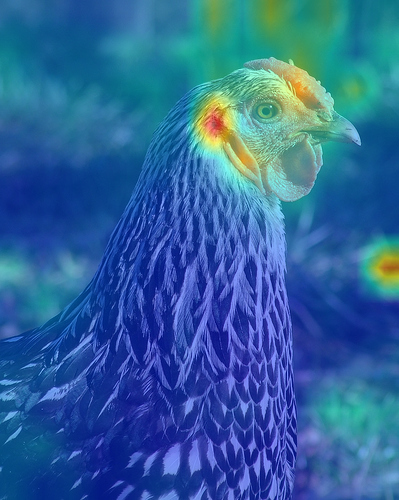}
        \end{minipage}
        \hfill
        \begin{minipage}[h]{0.09\linewidth}
            \includegraphics[width=\linewidth]{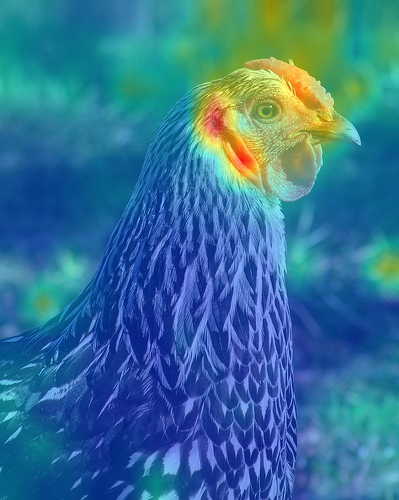}
        \end{minipage}
        \hfill
        \begin{minipage}[h]{0.09\linewidth}
            \includegraphics[width=\linewidth]{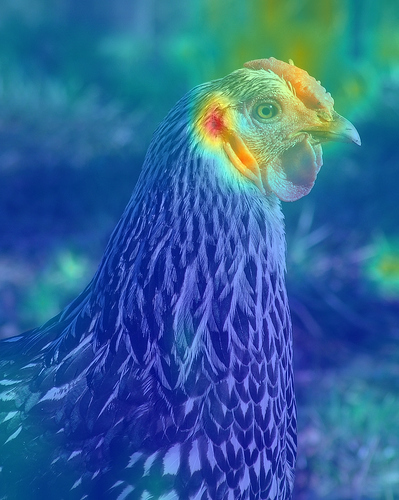}
        \end{minipage}
        \hfill
        \begin{minipage}[h]{0.09\linewidth}
            \includegraphics[width=\linewidth]{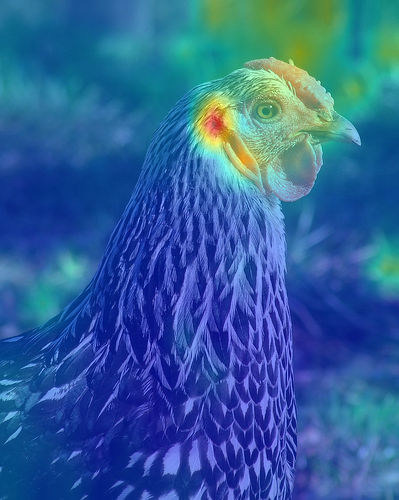}
        \end{minipage}
        \hfill
        \begin{minipage}[h]{0.09\linewidth}
            \includegraphics[width=\linewidth]{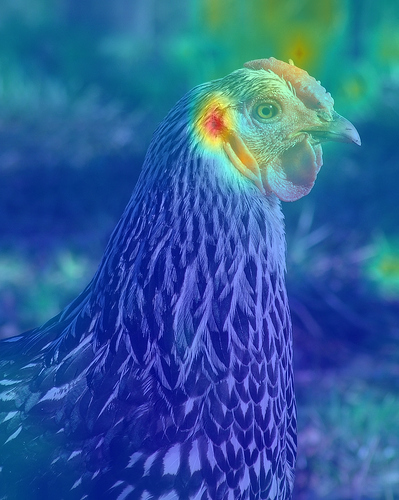}
        \end{minipage}
        \hfill
        \begin{minipage}[h]{0.09\linewidth}
            \includegraphics[width=\linewidth]{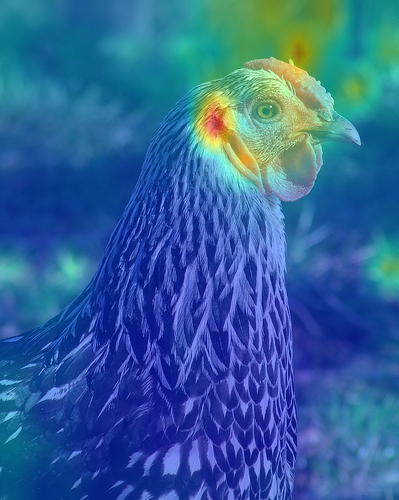}
        \end{minipage}
        \hfill
        \begin{minipage}[h]{0.09\linewidth}
            \includegraphics[width=\linewidth]{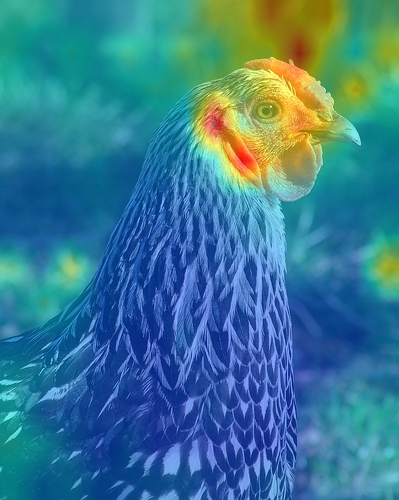}
        \end{minipage}
        \hfill
        \begin{minipage}[h]{0.09\linewidth}
            \includegraphics[width=\linewidth]{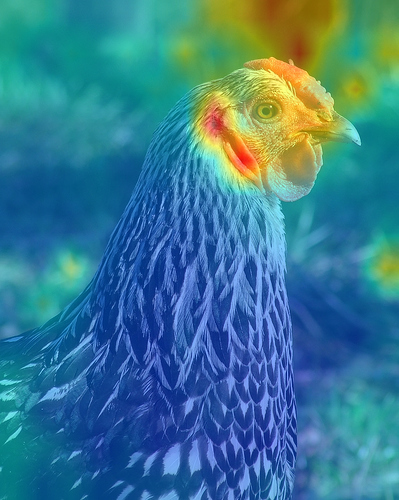}
        \end{minipage}
        \hfill
        \begin{minipage}[h]{0.09\linewidth}
            \includegraphics[width=\linewidth]{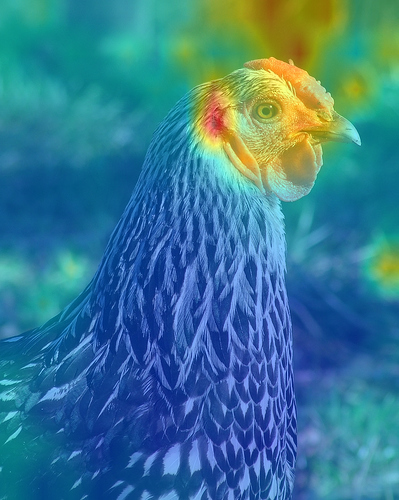}
        \end{minipage}
        \hfill
        \begin{minipage}[h]{0.09\linewidth}
            \includegraphics[width=\linewidth]{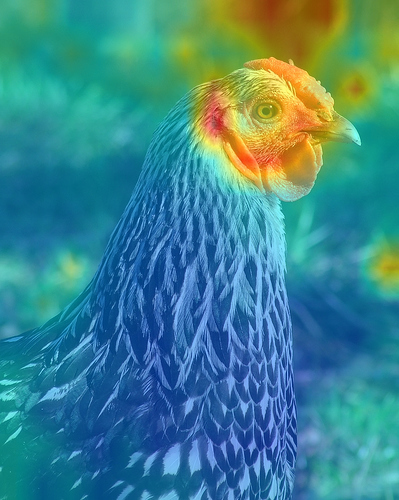}
        \end{minipage}

        \begin{minipage}[h]{0.09\linewidth}
            \includegraphics[width=\linewidth]{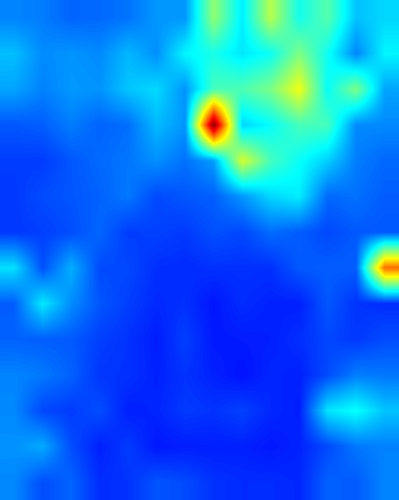}
        \end{minipage}
        \hfill
        \begin{minipage}[h]{0.09\linewidth}
            \includegraphics[width=\linewidth]{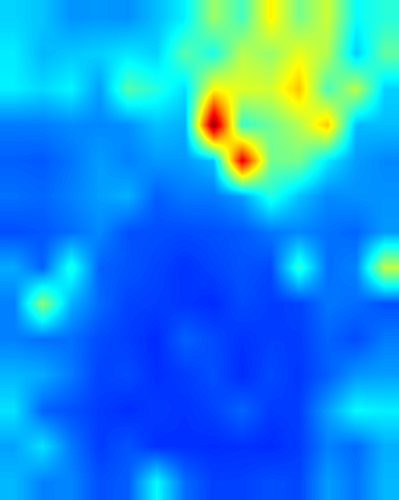}
        \end{minipage}
        \hfill
        \begin{minipage}[h]{0.09\linewidth}
            \includegraphics[width=\linewidth]{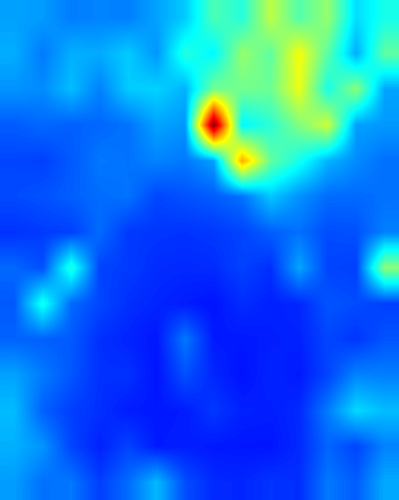}
        \end{minipage}
        \hfill
        \begin{minipage}[h]{0.09\linewidth}
            \includegraphics[width=\linewidth]{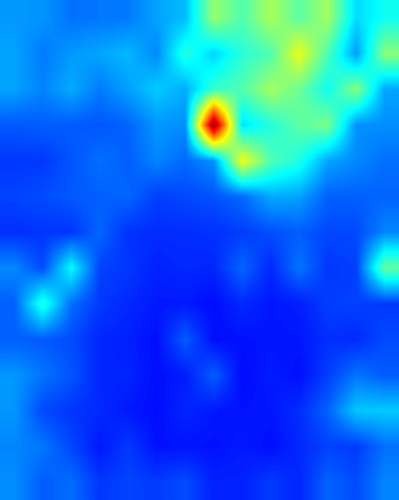}
        \end{minipage}
        \hfill
        \begin{minipage}[h]{0.09\linewidth}
            \includegraphics[width=\linewidth]{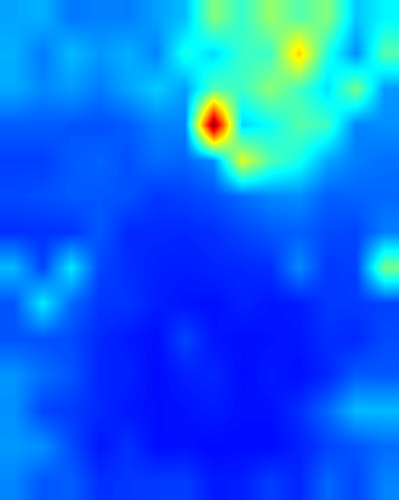}
        \end{minipage}
        \hfill
        \begin{minipage}[h]{0.09\linewidth}
            \includegraphics[width=\linewidth]{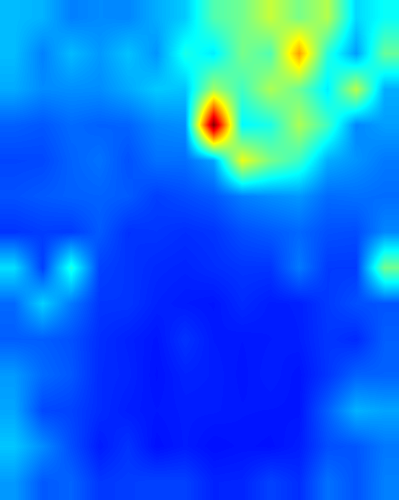}
        \end{minipage}
        \hfill
        \begin{minipage}[h]{0.09\linewidth}
            \includegraphics[width=\linewidth]{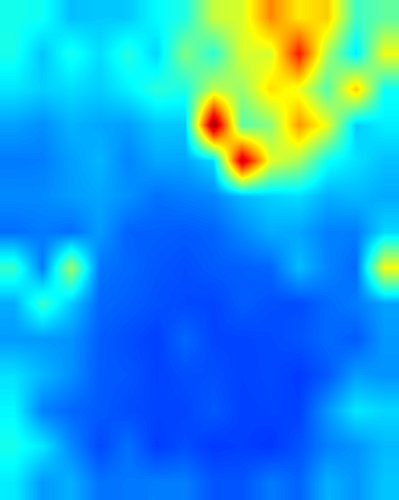}
        \end{minipage}
        \hfill
        \begin{minipage}[h]{0.09\linewidth}
            \includegraphics[width=\linewidth]{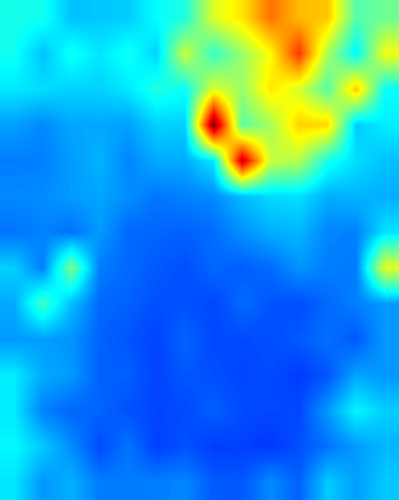}
        \end{minipage}
        \hfill
        \begin{minipage}[h]{0.09\linewidth}
            \includegraphics[width=\linewidth]{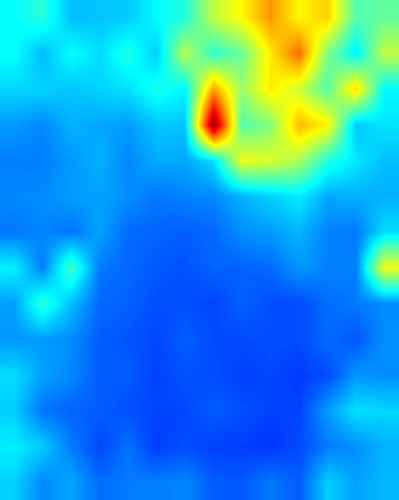}
        \end{minipage}
        \hfill
        \begin{minipage}[h]{0.09\linewidth}
            \includegraphics[width=\linewidth]{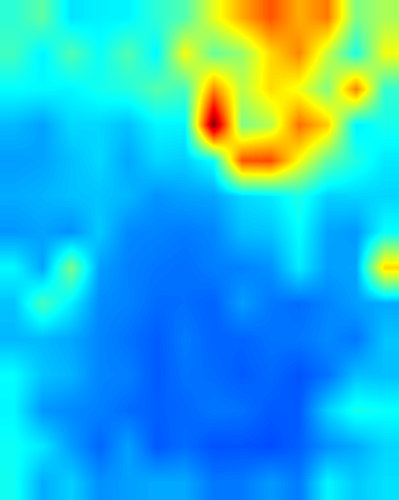}
        \end{minipage}
    }}

    \fbox{\parbox[c]{\linewidth}{
        Baseline w/LPA B10-10

        \begin{minipage}[h]{0.09\linewidth}
            \includegraphics[width=\linewidth]{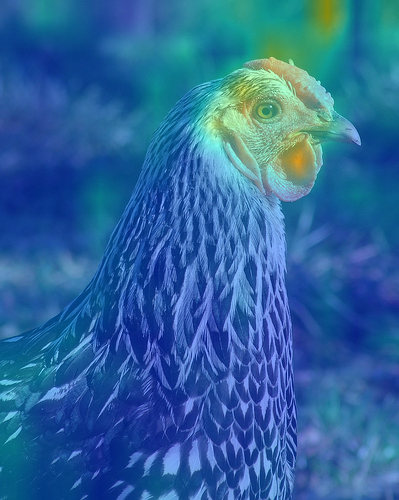}
        \end{minipage}
        \hfill
        \begin{minipage}[h]{0.09\linewidth}
            \includegraphics[width=\linewidth]{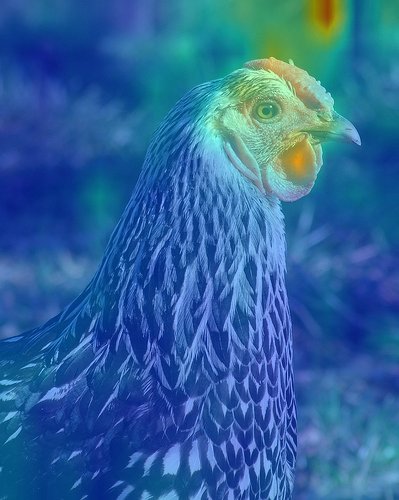}
        \end{minipage}
        \hfill
        \begin{minipage}[h]{0.09\linewidth}
            \includegraphics[width=\linewidth]{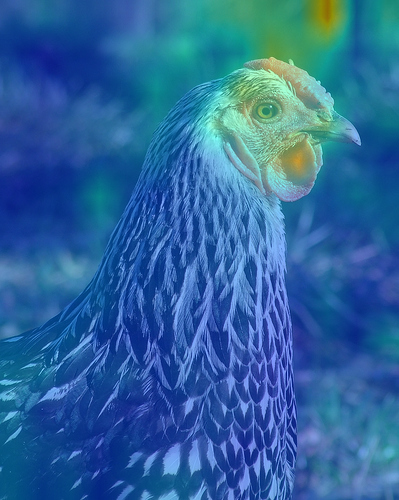}
        \end{minipage}
        \hfill
        \begin{minipage}[h]{0.09\linewidth}
            \includegraphics[width=\linewidth]{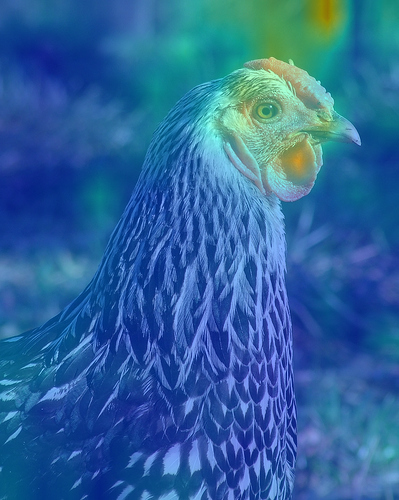}
        \end{minipage}
        \hfill
        \begin{minipage}[h]{0.09\linewidth}
            \includegraphics[width=\linewidth]{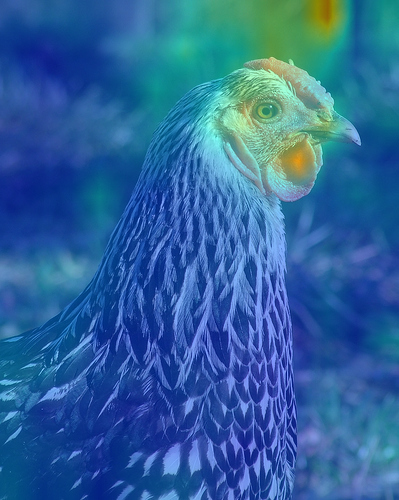}
        \end{minipage}
        \hfill
        \begin{minipage}[h]{0.09\linewidth}
            \includegraphics[width=\linewidth]{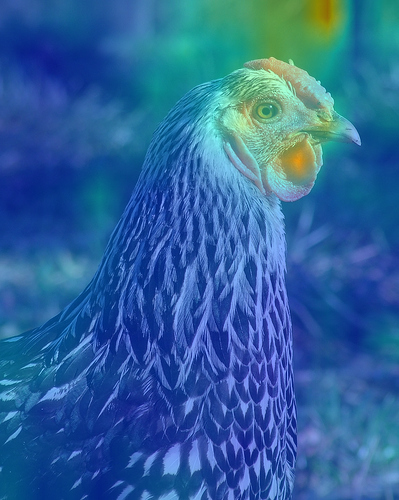}
        \end{minipage}
        \hfill
        \begin{minipage}[h]{0.09\linewidth}
            \includegraphics[width=\linewidth]{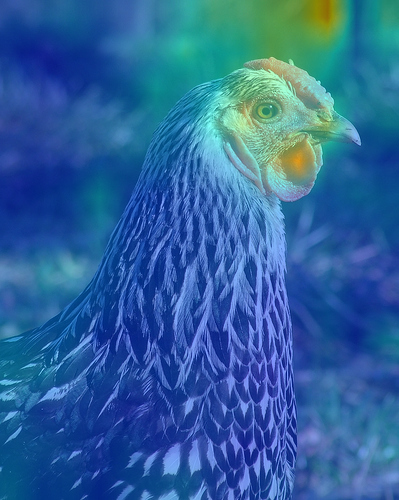}
        \end{minipage}
        \hfill
        \begin{minipage}[h]{0.09\linewidth}
            \includegraphics[width=\linewidth]{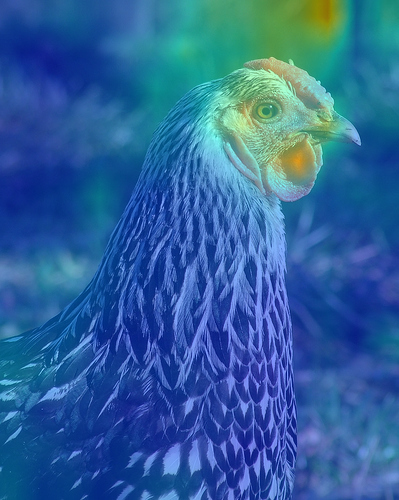}
        \end{minipage}
        \hfill
        \begin{minipage}[h]{0.09\linewidth}
            \includegraphics[width=\linewidth]{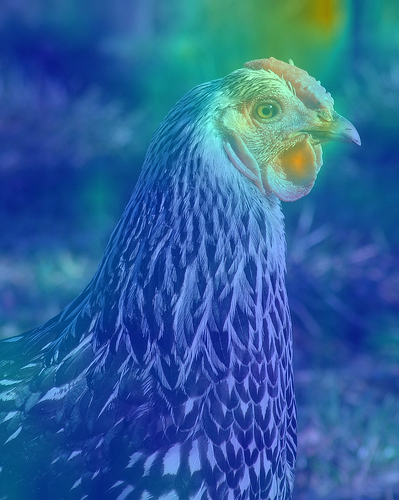}
        \end{minipage}
        \hfill
        \begin{minipage}[h]{0.09\linewidth}
            \includegraphics[width=\linewidth]{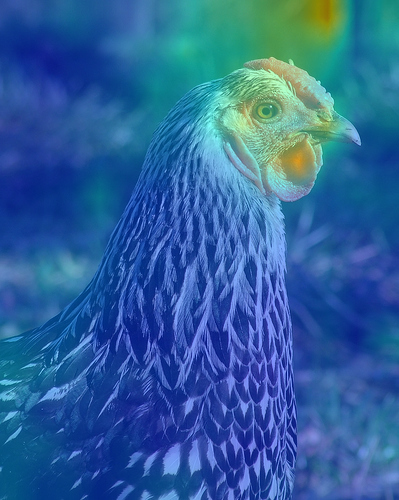}
        \end{minipage}

        \begin{minipage}[h]{0.09\linewidth}
            \includegraphics[width=\linewidth]{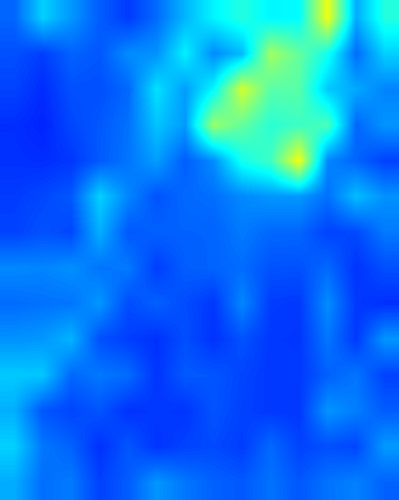}
        \end{minipage}
        \hfill
        \begin{minipage}[h]{0.09\linewidth}
            \includegraphics[width=\linewidth]{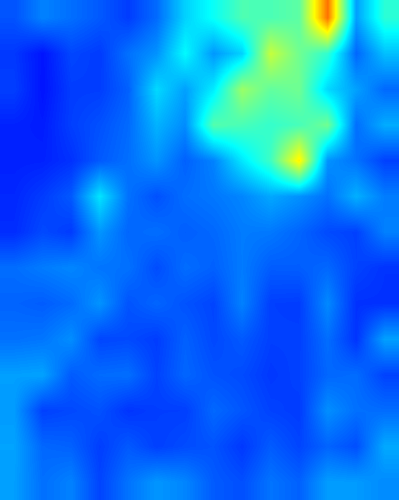}
        \end{minipage}
        \hfill
        \begin{minipage}[h]{0.09\linewidth}
            \includegraphics[width=\linewidth]{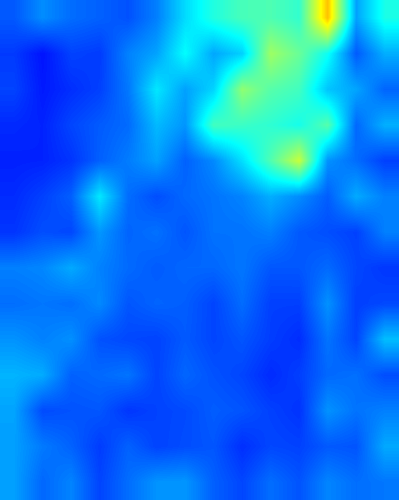}
        \end{minipage}
        \hfill
        \begin{minipage}[h]{0.09\linewidth}
            \includegraphics[width=\linewidth]{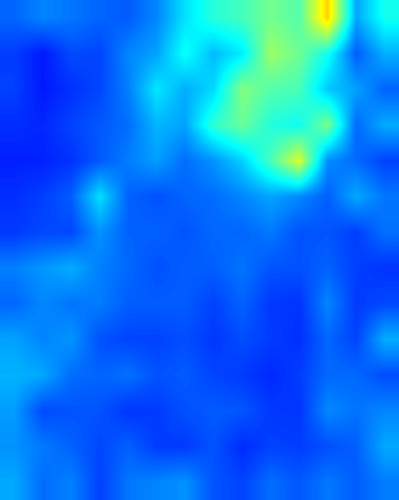}
        \end{minipage}
        \hfill
        \begin{minipage}[h]{0.09\linewidth}
            \includegraphics[width=\linewidth]{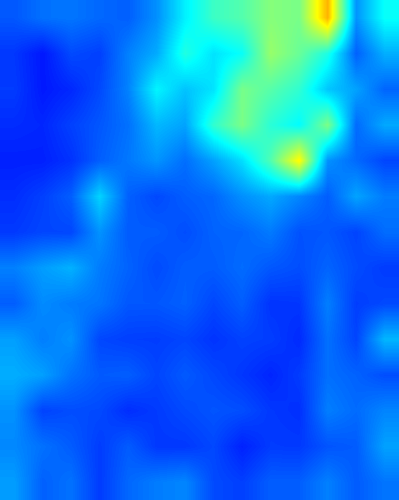}
        \end{minipage}
        \hfill
        \begin{minipage}[h]{0.09\linewidth}
            \includegraphics[width=\linewidth]{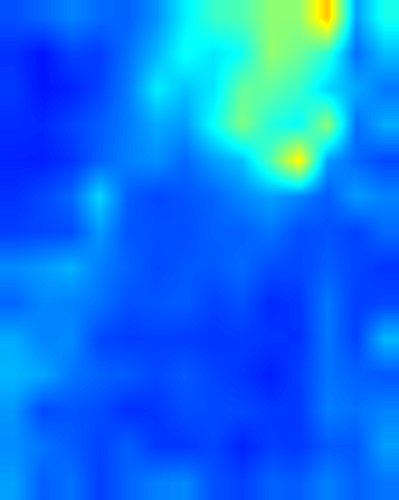}
        \end{minipage}
        \hfill
        \begin{minipage}[h]{0.09\linewidth}
            \includegraphics[width=\linewidth]{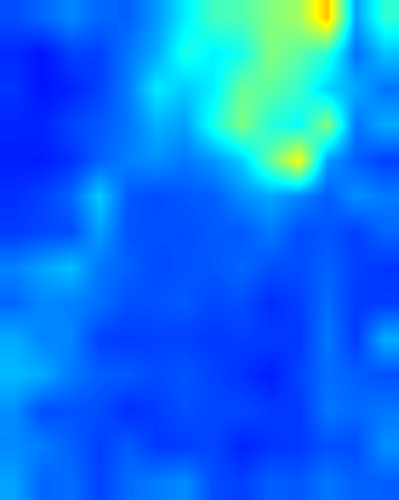}
        \end{minipage}
        \hfill
        \begin{minipage}[h]{0.09\linewidth}
            \includegraphics[width=\linewidth]{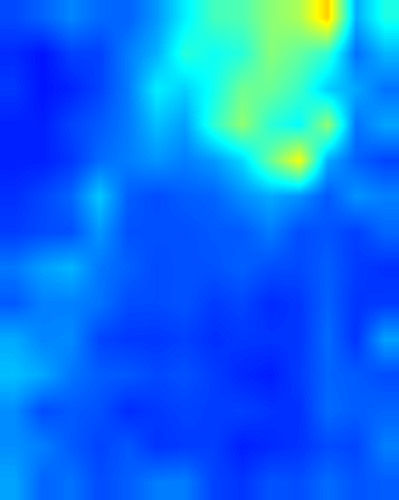}
        \end{minipage}
        \hfill
        \begin{minipage}[h]{0.09\linewidth}
            \includegraphics[width=\linewidth]{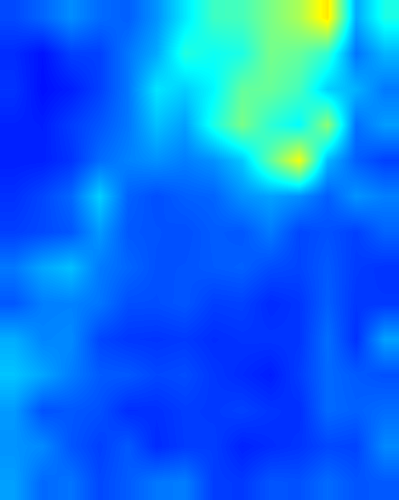}
        \end{minipage}
        \hfill
        \begin{minipage}[h]{0.09\linewidth}
            \includegraphics[width=\linewidth]{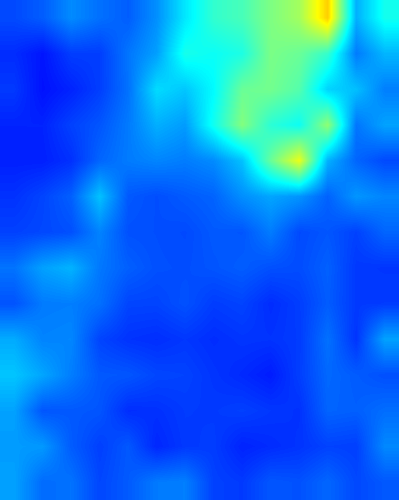}
        \end{minipage}
    }}

    \fbox{\parbox[c]{\linewidth}{
        Baseline B10-10

        \begin{minipage}[h]{0.09\linewidth}
            \includegraphics[width=\linewidth]{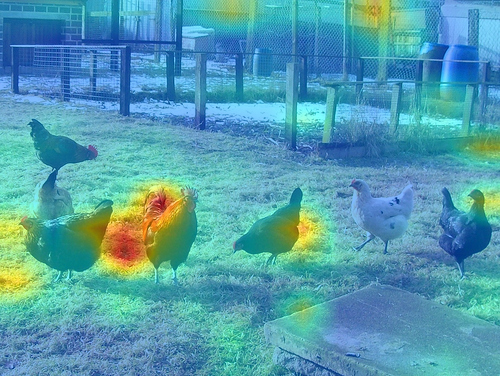}
        \end{minipage}
        \hfill
        \begin{minipage}[h]{0.09\linewidth}
            \includegraphics[width=\linewidth]{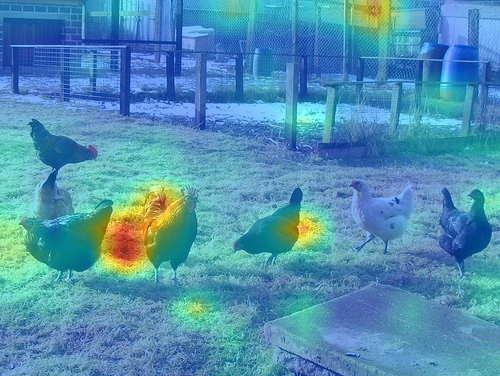}
        \end{minipage}
        \hfill
        \begin{minipage}[h]{0.09\linewidth}
            \includegraphics[width=\linewidth]{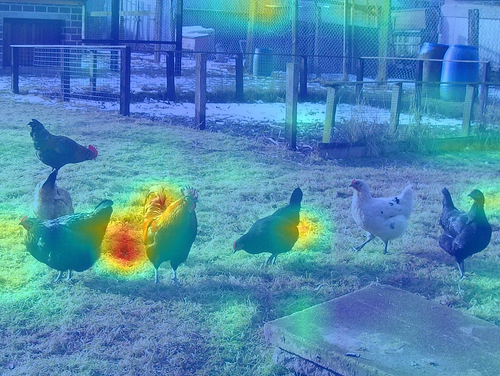}
        \end{minipage}
        \hfill
        \begin{minipage}[h]{0.09\linewidth}
            \includegraphics[width=\linewidth]{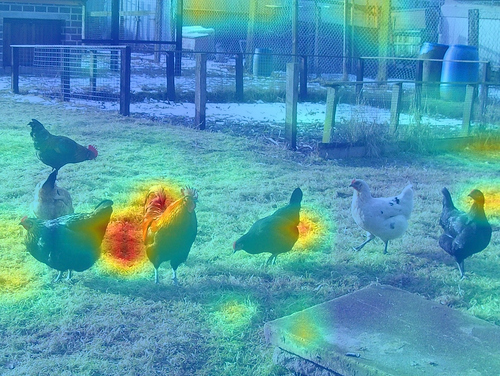}
        \end{minipage}
        \hfill
        \begin{minipage}[h]{0.09\linewidth}
            \includegraphics[width=\linewidth]{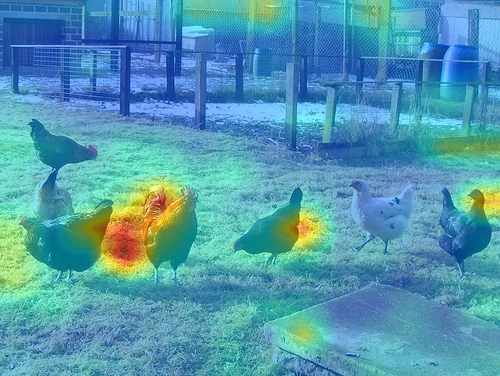}
        \end{minipage}
        \hfill
        \begin{minipage}[h]{0.09\linewidth}
            \includegraphics[width=\linewidth]{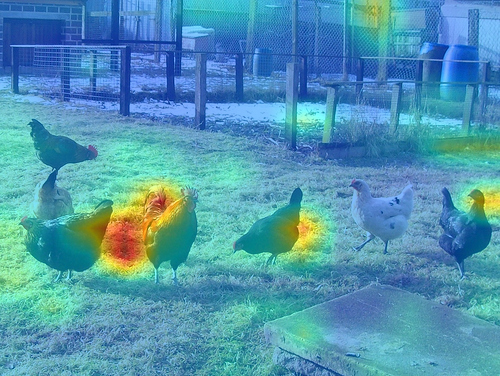}
        \end{minipage}
        \hfill
        \begin{minipage}[h]{0.09\linewidth}
            \includegraphics[width=\linewidth]{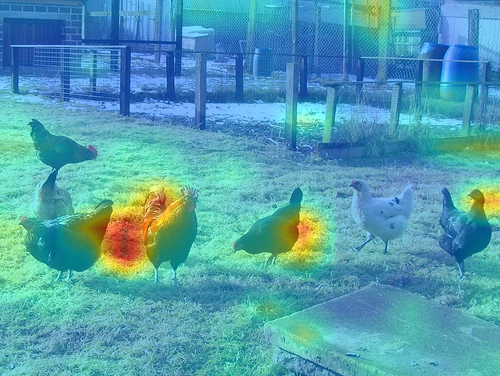}
        \end{minipage}
        \hfill
        \begin{minipage}[h]{0.09\linewidth}
            \includegraphics[width=\linewidth]{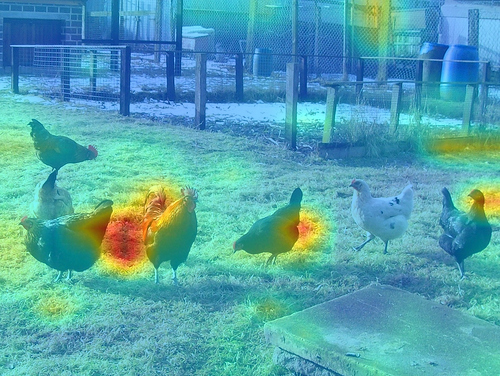}
        \end{minipage}
        \hfill
        \begin{minipage}[h]{0.09\linewidth}
            \includegraphics[width=\linewidth]{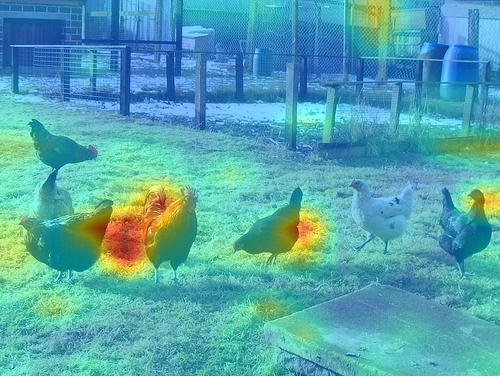}
        \end{minipage}
        \hfill
        \begin{minipage}[h]{0.09\linewidth}
            \includegraphics[width=\linewidth]{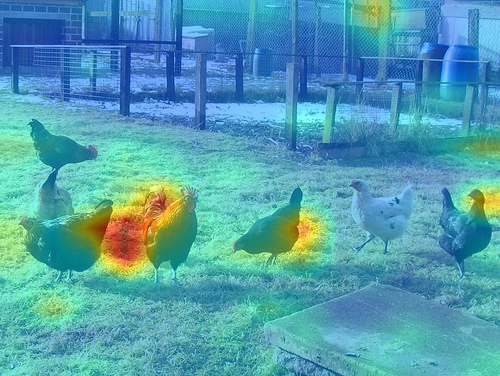}
        \end{minipage}

        \begin{minipage}[h]{0.09\linewidth}
            \includegraphics[width=\linewidth]{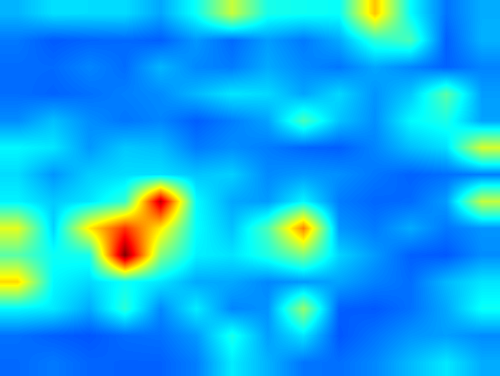}
        \end{minipage}
        \hfill
        \begin{minipage}[h]{0.09\linewidth}
            \includegraphics[width=\linewidth]{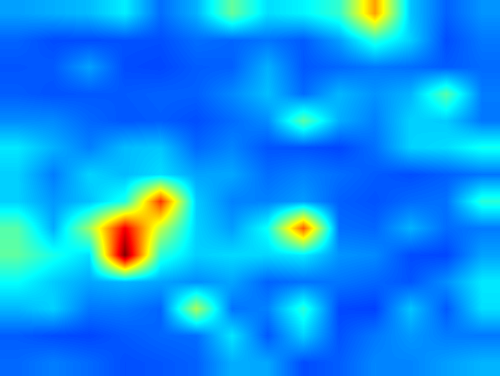}
        \end{minipage}
        \hfill
        \begin{minipage}[h]{0.09\linewidth}
            \includegraphics[width=\linewidth]{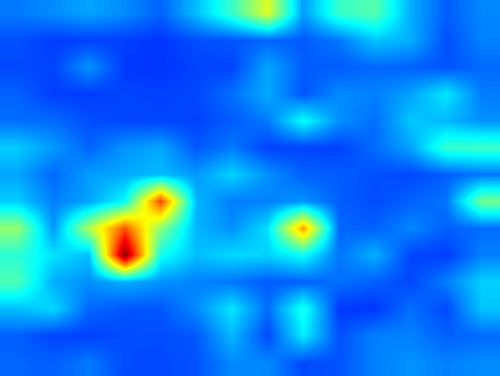}
        \end{minipage}
        \hfill
        \begin{minipage}[h]{0.09\linewidth}
            \includegraphics[width=\linewidth]{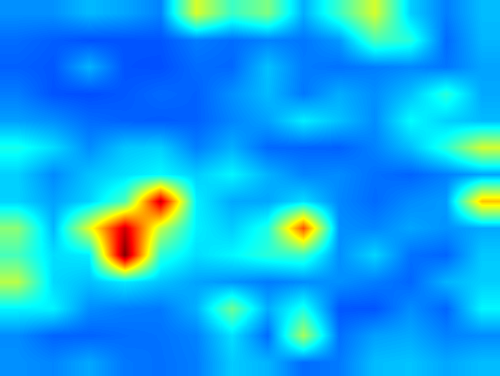}
        \end{minipage}
        \hfill
        \begin{minipage}[h]{0.09\linewidth}
            \includegraphics[width=\linewidth]{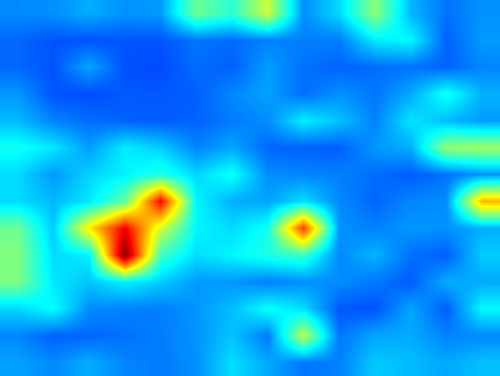}
        \end{minipage}
        \hfill
        \begin{minipage}[h]{0.09\linewidth}
            \includegraphics[width=\linewidth]{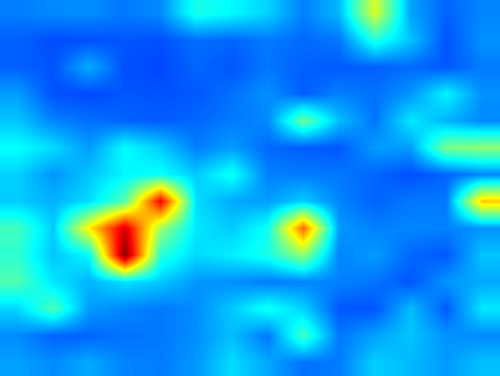}
        \end{minipage}
        \hfill
        \begin{minipage}[h]{0.09\linewidth}
            \includegraphics[width=\linewidth]{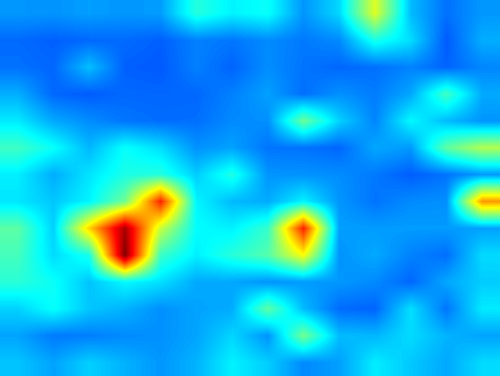}
        \end{minipage}
        \hfill
        \begin{minipage}[h]{0.09\linewidth}
            \includegraphics[width=\linewidth]{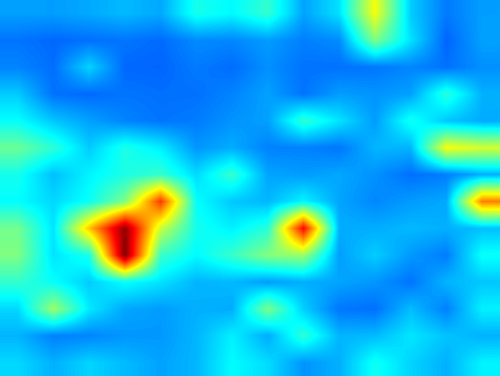}
        \end{minipage}
        \hfill
        \begin{minipage}[h]{0.09\linewidth}
            \includegraphics[width=\linewidth]{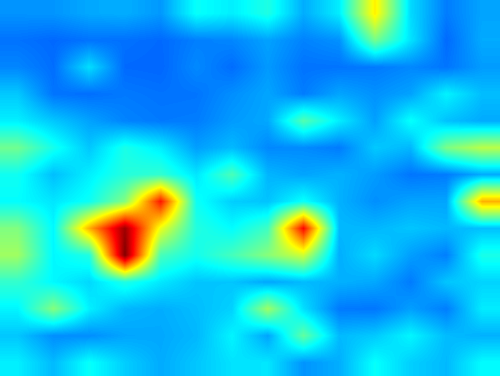}
        \end{minipage}
        \hfill
        \begin{minipage}[h]{0.09\linewidth}
            \includegraphics[width=\linewidth]{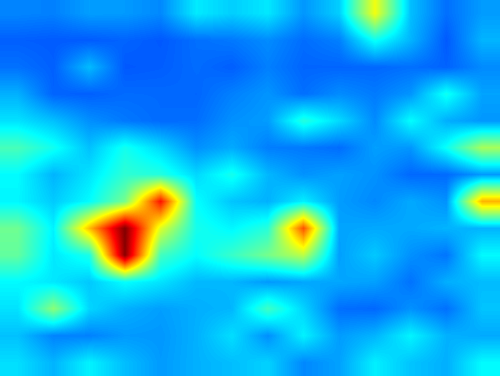}
        \end{minipage}
    }}

    \fbox{\parbox[c]{\linewidth}{
        Baseline w/LPA B10-10

        \begin{minipage}[h]{0.09\linewidth}
            \includegraphics[width=\linewidth]{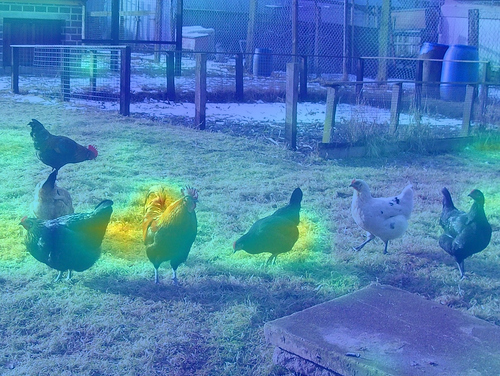}
        \end{minipage}
        \hfill
        \begin{minipage}[h]{0.09\linewidth}
            \includegraphics[width=\linewidth]{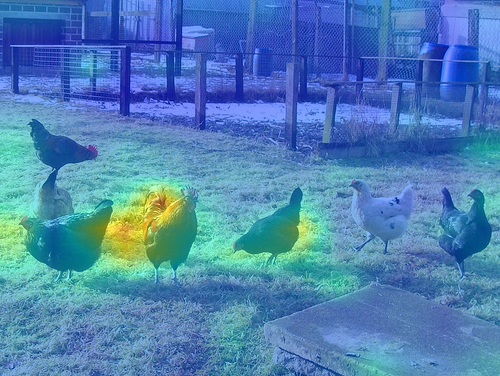}
        \end{minipage}
        \hfill
        \begin{minipage}[h]{0.09\linewidth}
            \includegraphics[width=\linewidth]{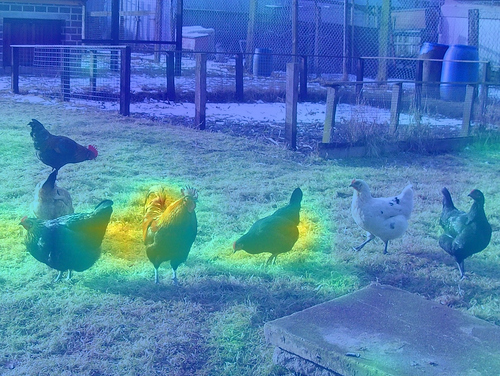}
        \end{minipage}
        \hfill
        \begin{minipage}[h]{0.09\linewidth}
            \includegraphics[width=\linewidth]{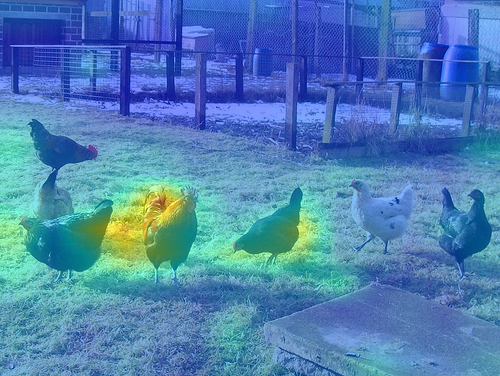}
        \end{minipage}
        \hfill
        \begin{minipage}[h]{0.09\linewidth}
            \includegraphics[width=\linewidth]{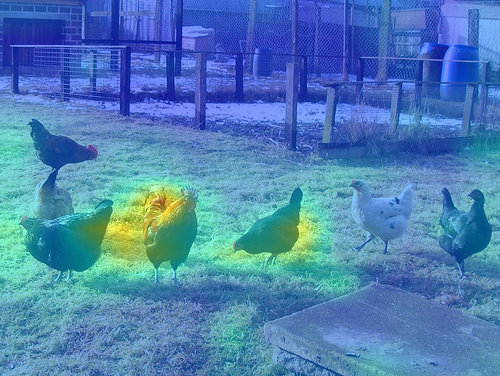}
        \end{minipage}
        \hfill
        \begin{minipage}[h]{0.09\linewidth}
            \includegraphics[width=\linewidth]{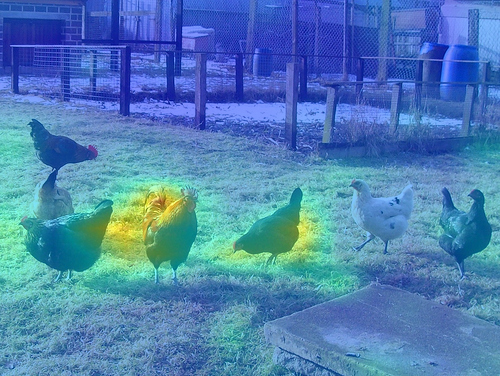}
        \end{minipage}
        \hfill
        \begin{minipage}[h]{0.09\linewidth}
            \includegraphics[width=\linewidth]{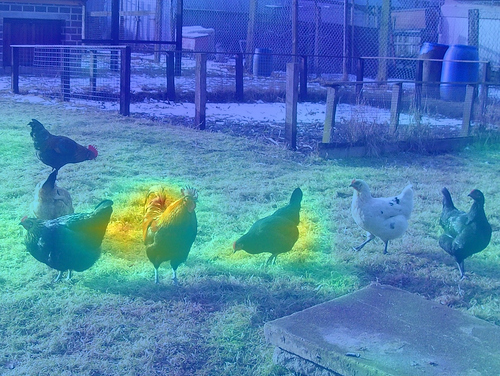}
        \end{minipage}
        \hfill
        \begin{minipage}[h]{0.09\linewidth}
            \includegraphics[width=\linewidth]{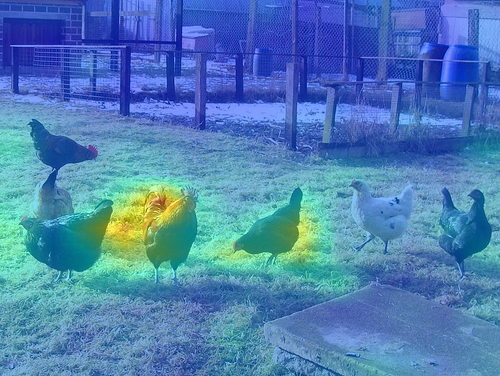}
        \end{minipage}
        \hfill
        \begin{minipage}[h]{0.09\linewidth}
            \includegraphics[width=\linewidth]{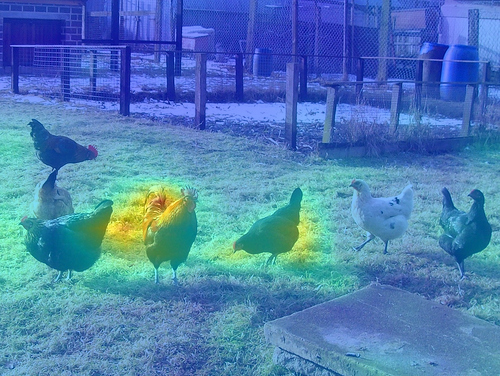}
        \end{minipage}
        \hfill
        \begin{minipage}[h]{0.09\linewidth}
            \includegraphics[width=\linewidth]{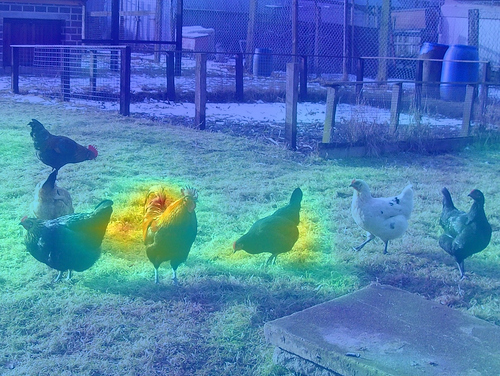}
        \end{minipage}

        \begin{minipage}[h]{0.09\linewidth}
            \includegraphics[width=\linewidth]{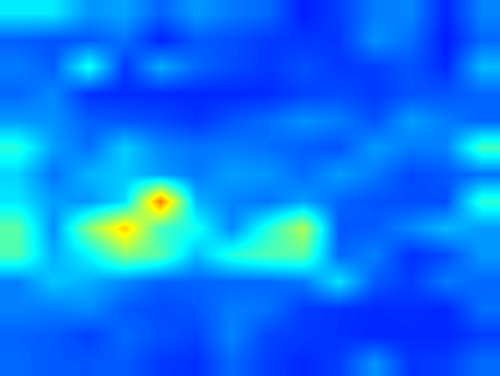}
        \end{minipage}
        \hfill
        \begin{minipage}[h]{0.09\linewidth}
            \includegraphics[width=\linewidth]{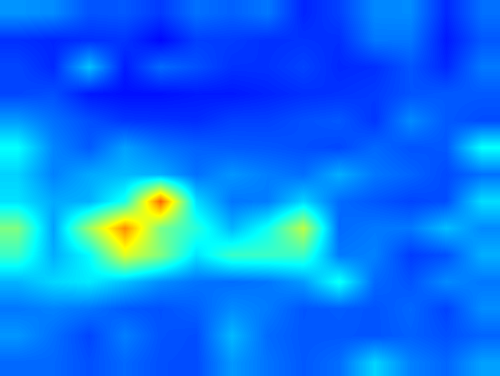}
        \end{minipage}
        \hfill
        \begin{minipage}[h]{0.09\linewidth}
            \includegraphics[width=\linewidth]{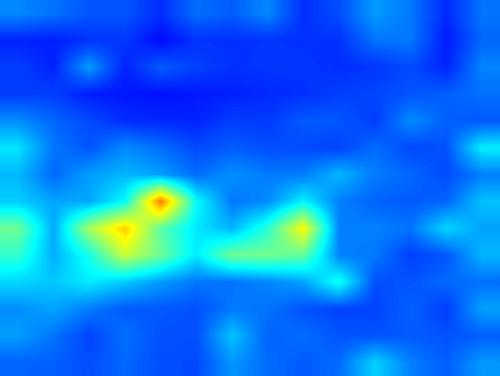}
        \end{minipage}
        \hfill
        \begin{minipage}[h]{0.09\linewidth}
            \includegraphics[width=\linewidth]{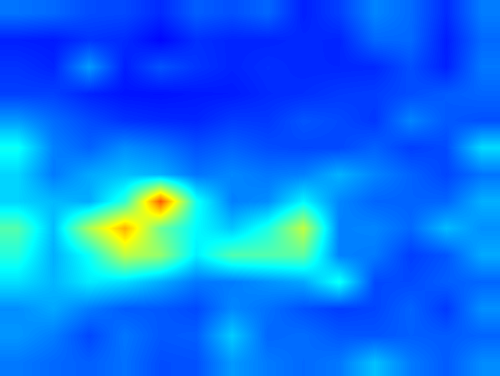}
        \end{minipage}
        \hfill
        \begin{minipage}[h]{0.09\linewidth}
            \includegraphics[width=\linewidth]{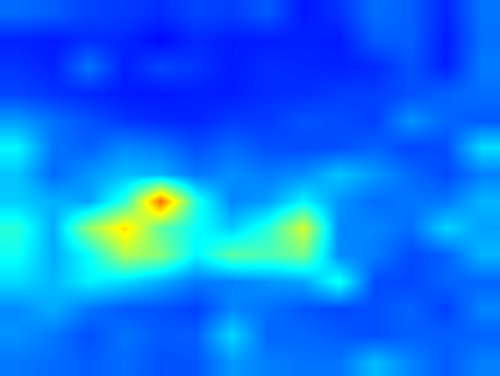}
        \end{minipage}
        \hfill
        \begin{minipage}[h]{0.09\linewidth}
            \includegraphics[width=\linewidth]{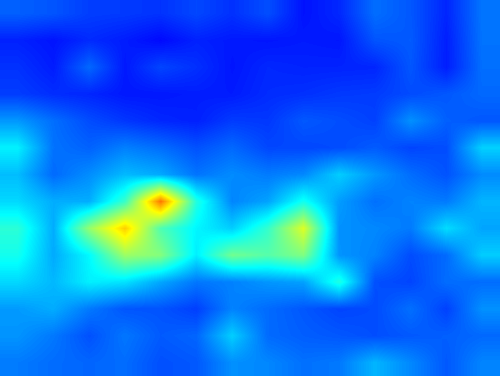}
        \end{minipage}
        \hfill
        \begin{minipage}[h]{0.09\linewidth}
            \includegraphics[width=\linewidth]{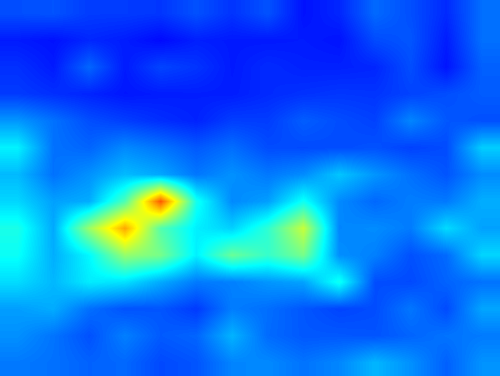}
        \end{minipage}
        \hfill
        \begin{minipage}[h]{0.09\linewidth}
            \includegraphics[width=\linewidth]{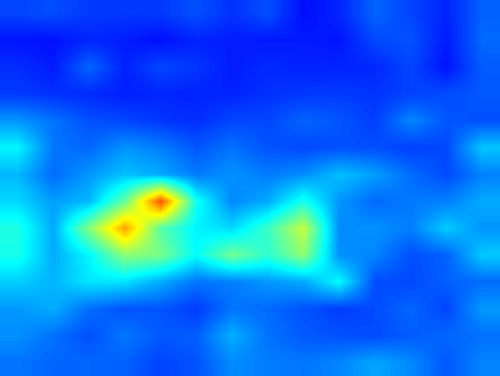}
        \end{minipage}
        \hfill
        \begin{minipage}[h]{0.09\linewidth}
            \includegraphics[width=\linewidth]{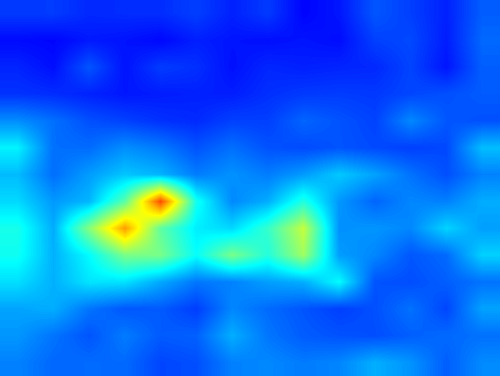}
        \end{minipage}
        \hfill
        \begin{minipage}[h]{0.09\linewidth}
            \includegraphics[width=\linewidth]{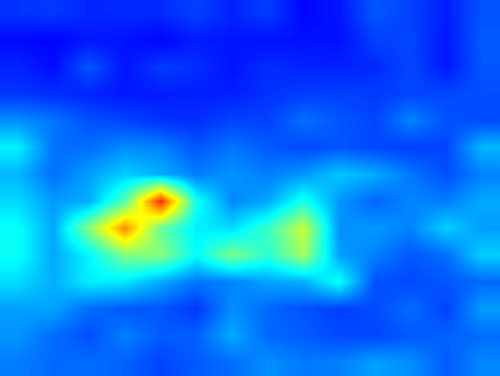}
        \end{minipage}
    }}
    \caption{Comparison of Attention Map Evolutions}
    \label{fig:attevol}
\end{figure}

\textit{Attention Rollout}~\cite{abnar2020quantifying} is a simple technique to analyze the attention flow for ViT models. 
To compute the attention flow from layer $j$ to $i$:
\begin{equation}
    \tilde{A}^{(i)}=\begin{cases}
        A^{(i)}\tilde{A}^{(i-1)}&, i>j\\
        A^{(i)}&, i=j
    \end{cases},
\end{equation}
where $A^{(i)}$ is the attention matrix of layer $i$. 
We take the attention of the class token for attention map visualization.
We provide more attention map evolution results in Figure~\ref{fig:attevol}. 
As we can see, the attention maps of the baseline focus more on the irrelevant part of the image than the baseline with LPA.

\end{document}